\documentclass[runningheads]{llncs}

\usepackage{eccv}

\usepackage{eccvabbrv}

\usepackage{graphicx}
\usepackage{booktabs}
\usepackage{amsmath}
\usepackage{amssymb}
\usepackage{amsfonts}
\usepackage{mathtools}
\usepackage{algorithm}
\usepackage{algorithmic}
\usepackage{multirow}
\usepackage{xcolor}
\usepackage{colortbl}
\usepackage{float}
\usepackage{pifont}
\usepackage{caption}
\usepackage{subcaption}

\newcommand{\cmark}{\checkmark}
\newcommand{\xmark}{\ding{55}}

\definecolor{darkgreen}{rgb}{0.0, 0.5, 0.0}
\definecolor{text_red}{RGB}{220,20,60}
\definecolor{DeepPurple}{RGB}{112, 48, 140}
\definecolor{DeepRed}{RGB}{192, 0, 0}

\usepackage[accsupp]{axessibility}

\usepackage{hyperref}

\usepackage{orcidlink}

\begin{document}

\title{TARS: MinMax Token-Adaptive Preference Strategy for Hallucination Reduction in MLLMs}

\titlerunning{TARS}

\author{Kejia Zhang\textsuperscript{1} \and
Keda Tao\textsuperscript{2} \and
Zhiming Luo\textsuperscript{1*} \and
Chang Liu\textsuperscript{4$\dagger$} \and
Jiasheng Tang\textsuperscript{3,5} \and
Huan Wang\textsuperscript{2*}}

\authorrunning{K.~Zhang et al.}

\institute{\textsuperscript{1}Xiamen University \quad \textsuperscript{2}Westlake University \quad \textsuperscript{3}DAMO Academy, Alibaba Group\\
\textsuperscript{4}AWS AI Lab, Amazon \quad \textsuperscript{5}Hupan Laboratory\\
\textsuperscript{*}Corresponding Author.\\
\textsuperscript{$\dagger$}Work done prior to joining Amazon.}

\maketitle

\vspace{-5mm}
\begin{center}
    \begin{tabular}{@{}c@{\hspace{2mm}}c@{}}
        \includegraphics[width=0.54\linewidth]{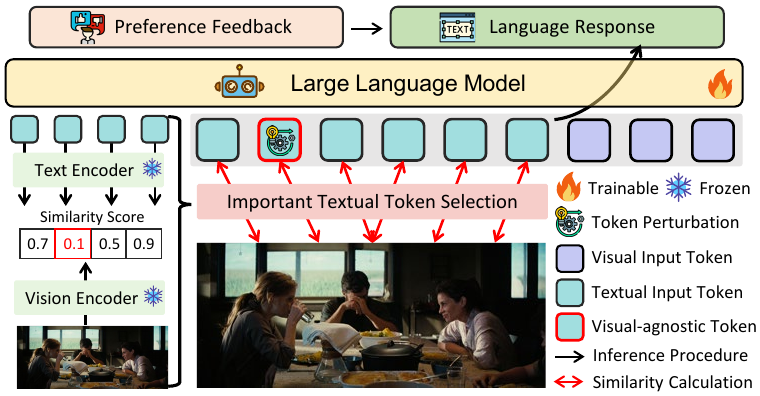} &
        \raisebox{-0.1cm}{\includegraphics[width=0.44\linewidth,height=0.3\linewidth]{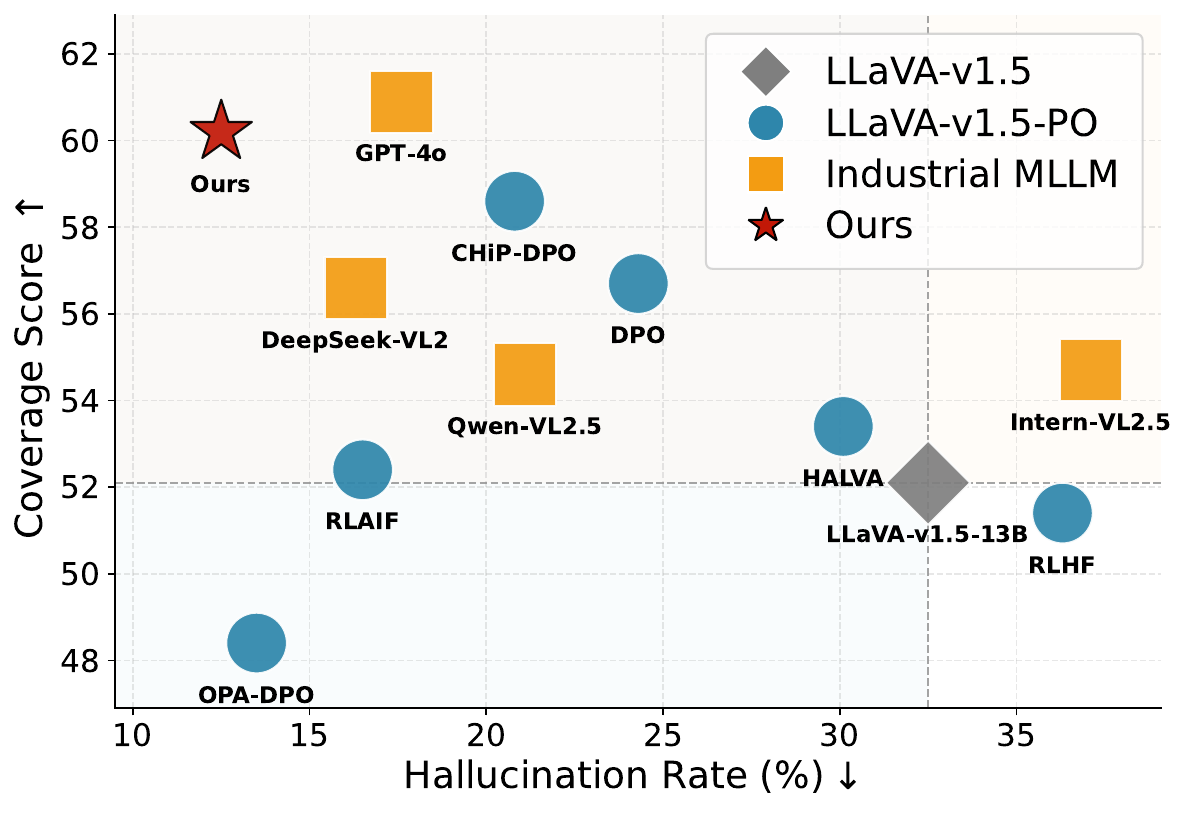}}
    \end{tabular}
    \vspace{-2mm}
    \captionof{figure}{
    \textbf{Left:} We present \textit{TARS}, a \underline{t}oken-\underline{a}daptive p\underline{r}eference \underline{s}trategy for mitigating hallucinations in MLLMs.
    TARS reformulates direct preference optimization (DPO) as a principled min-max optimization objective:
    (1) minimizes behavioral misalignment via structured preference feedback supervision and
    (2) maximizes distributional adaptability through controlled perturbations of visual-agnostic tokens.
    \textbf{Right:} Comprehensive evaluation on LLaVA-v1.5-13B with preference optimization (PO)~\cite{llava_origin} and various state-of-the-art MLLMs under the AMBER benchmark~\cite{wang2023amber} shows that TARS consistently surpasses PO baselines, yielding results competitive with GPT-4o~\cite{hurst2024gpt}.}
    \label{fig:teaser}
\end{center}
\vspace{-5mm}

\begin{abstract}
Multimodal large language models (MLLMs) are prone to hallucinations, generating plausible but visually ungrounded outputs, partly because direct preference optimization (DPO) overfits to superficial linguistic cues under static preference supervision.
We propose TARS, a token-adaptive preference strategy that reformulates DPO as a principled min-max optimization problem. The inner maximization selectively perturbs visual-agnostic tokens to induce worst-case distributional shifts, while the outer minimization enforces alignment with causal visual signals rather than surface-level patterns. A novel spectral alignment loss further regularizes hidden representations in the frequency domain via the Fast Fourier Transform (FFT), preserving global semantic structure without rigid token-level correspondence.
We evaluate TARS across multiple hallucination benchmarks. Using only 4.8k preference samples without expert feedback, TARS reduces hallucination rates from 26.4\% to 13.2\% and cognition scores from 2.5 to 0.4, outperforming standard DPO by a large margin. Notably, TARS surpasses $5\times$ LLM-based data augmentation trained on 28.8k samples (Hal-Rate: 16.0\% vs.\ 13.2\%), demonstrating that reshaping the optimization landscape via adversarial token perturbation is fundamentally more effective than scaling training data. TARS further narrows the gap with GPT-4o on key metrics.
\end{abstract}

\section{Introduction}
\label{sec:intro}
Multimodal large language models (MLLMs) extend the reasoning capabilities of LLMs~\cite{gandhi2023understanding, chen2023large, wang2025reasoning} to visual inputs, enabling grounded vision-language understanding~\cite{tong2024cambrian, huang2023language, feng2025can}. Despite strong performance across diverse tasks~\cite{jiang2024marvel, shao2025holitom}, MLLMs remain prone to hallucinations, producing outputs that appear plausible but are factually incorrect or lack visual grounding~\cite{kim2024code, jiang2024hallucination, huang2024opera, sarkarmitigating, gunjal2024detecting}. Mitigating such failures is crucial for deploying reliable MLLMs.

\begin{figure}[tb]
        \begin{center}
        \includegraphics[width=\linewidth]{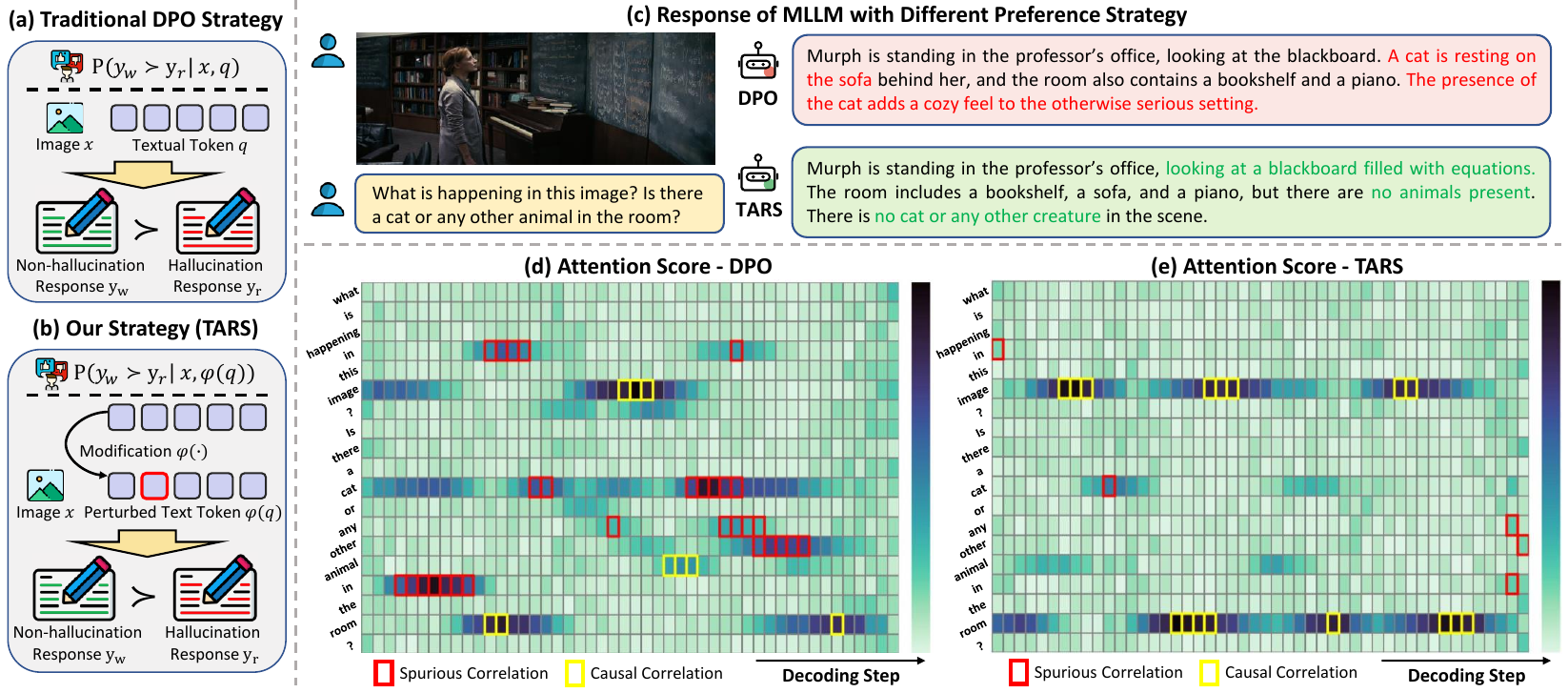}
        \end{center}
        \vspace{-1.5em}
        \caption{Motivation illustration for TARS.
        (a) and (b) illustrate standard DPO and our token-adaptive perturbation strategy.
        (c) shows a VQA example where DPO hallucinates, while TARS effectively avoids ungrounded output.
        (d) and (e) visualize token-to-query attention maps during autoregressive decoding. DPO over-attends to spurious tokens, while TARS attends to causally grounded visual-semantic cues.
        }
        \label{fig:claim}
        \vspace{-1.5em}
\end{figure}

Modern MLLMs follow a two-stage pipeline of knowledge pretraining~\cite{instruct_blip, bao2022vlmo, 10445007} and instruction tuning~\cite{liu2024improved, liu2023aligning}. Hallucinations often stem not from knowledge deficits but from behavioral biases that produce plausible yet ungrounded outputs~\cite{oh2024erbench, chen2025perturbollava}. Preference optimization (PO) addresses this by fine-tuning with ranked response pairs from human~\cite{RLHF, RLHF_1} or AI feedback~\cite{RLAIF, sharma2024critical}, aligning outputs with factual expectations~\cite{PPO, achiam2017constrained}.
Direct preference optimization~(DPO)~\cite{rafailov2023direct} is widely used for hallucination reduction~\cite{CHIP, OPA_DPO}, yet current methods can overfit to shallow textual cues such as high-frequency phrases~\cite{huang2024opera, liu2024paying}, generating plausible but visually ungrounded responses (\cref{fig:claim}(c)). Our analysis (\cref{fig:claim}(d)) further reveals that DPO-trained models assign high preference to outputs with spurious correlation tokens (\eg, prepositions or frequently mentioned objects) that lack visual grounding~\cite{xie2024v, wang2024mdpo}. This reliance on static preference signals hinders generalization under shifting visual-textual contexts, leading to brittle alignment and increased hallucination~\cite{setlur2024rl, CHIP}.

We formulate this challenge as a \textbf{min-max token-adaptive alignment problem}: maximizing distributional variation under semantic constraints, then minimizing preference loss under these perturbations.
A natural alternative is data augmentation, which diversifies the preference distribution by generating additional training pairs.
However, our experiments (\cref{sec:augmentation_compare}) reveal that even $5\times$ LLM-based augmentation (28.8k samples) underperforms TARS using only the original 4.8k samples (Hal-Rate: 16.0\% vs.\ 13.2\%), demonstrating that reshaping the optimization landscape via adversarial token perturbation is fundamentally more effective than merely scaling training data. The key distinction is that data augmentation merely enriches the \textbf{data distribution}, whereas TARS actively reshapes the \textbf{optimization landscape}: the inner maximization dynamically generates worst-case token configurations per sample, forcing the outer minimization to learn alignment features that are invariant to distributional shifts rather than memorizing surface-level patterns.

Specifically, we perturb visual-agnostic tokens, \ie, textual elements with minimal cross-modal grounding, to shift the input distribution without altering semantic content. This forces the model to rely on causally grounded visual signals rather than superficial correlations (\cref{fig:claim}(b)).
Supervising perturbed representations requires maintaining semantic consistency without rigid token-level correspondence. Standard spatial constraints (\eg, $\ell_2$ or contrastive losses) indiscriminately penalize any positional deviation equally, reintroducing the spurious correlations our perturbation aims to eliminate~\cite{zhou2024explore, tianblack}.
We instead propose a novel \textbf{spectral alignment loss} via the Fast Fourier Transform (FFT). Under the shift property of the Discrete Fourier Transform, a local token perturbation at position $l$ contributes a bounded additive term proportional to $e^{-2\pi i k l / L}$ to each frequency bin $k$, regardless of $l$~\cite{cooley1965algorithm}. This means spectral magnitudes vary smoothly under local perturbation, whereas spatial metrics ($\ell_2$, cosine) exhibit sharp discontinuities at the perturbed positions. Meanwhile, the dominant low-frequency components capture global semantic structure (\eg, topic, intent, visual grounding), while high-frequency components encode position-dependent lexical details. This makes frequency-domain alignment a theoretically principled choice for preserving global semantic invariance while tolerating the local distributional shifts inherent in our min-max formulation (\cref{sec:spectral}).\footnote{Empirical comparisons between spectral and spatial alignment strategies are provided in the Appendix, Section~9.4.}

We refer to this approach as \textbf{TARS} (\underline{t}oken-\underline{a}daptive p\underline{r}eference \underline{s}trategy), a lightweight and generalizable approach that enhances preference learning by combining adaptive token perturbation with frequency-domain spectral regularization.
We evaluate TARS on LLaVA-v1.5~\cite{llava_origin} at 7B and 13B scales across generative and discriminative hallucination benchmarks. TARS achieves consistently strong performance across all benchmarks and matches GPT-4o~\cite{hurst2024gpt} in several settings, underscoring the effectiveness of token-adaptive preference optimization.
Our contributions are as follows:
\begin{itemize}
    \item We reformulate preference learning as a min-max optimization that dynamically generates worst-case token perturbations per sample, reshaping the optimization landscape rather than augmenting data. With only 4.8k samples, TARS outperforms $5\times$ LLM-based augmentation (28.8k samples).
    \item We propose a \textbf{spectral alignment loss} via FFT that preserves global semantic structure in dominant low-frequency components while tolerating local distributional shifts from token perturbation, avoiding spurious positional correlations reintroduced by spatial constraints (\eg, $\ell_2$, cosine).
    \item We present \textbf{TARS}, combining adaptive token perturbation with spectral regularization. Using only 4.8k samples without expert feedback, TARS achieves state-of-the-art hallucination reduction, matching or surpassing GPT-4o on key metrics while preserving general reasoning capabilities.
\end{itemize}

\section{Preliminaries}
\label{sec:prelim}

\noindent\textbf{Multimodal Large Language Models.}
MLLMs extend LLMs by incorporating visual inputs alongside textual prompts~\cite{10445007}. Formally, given an image $x$ and a prompt $q$, the model generates a textual response $y = (y_1, \dots, y_l)$ in an autoregressive manner~\cite{llava_origin}:
\begin{equation}
    y_t \sim \pi_\theta(y_t \mid y_{<t}, x, q),
\end{equation}
where $\pi_\theta$ denotes the conditional generation policy parameterized by $\theta$. Given a textual input $q$ and a visual input $x$, the model tokenizes them into discrete sequences: textual tokens $q = \{q_1, \dots, q_m\}$ and visual tokens $x = \{x_1, \dots, x_n\}$. In practice, a pretrained vision encoder extracts patch-level features from the image, which are projected into the language model's embedding space through a learnable alignment module. These tokens are mapped to embeddings and fused via cross-attention to integrate semantic signals from both modalities. The resulting multimodal context is then used by the decoder to autoregressively generate the output sequence~\cite{dou2022empirical, yang2021causal}.

\noindent\textbf{Direct Preference Optimization.}
Direct preference optimization (DPO)~\cite{rafailov2023direct} is an effective and widely adopted approach for aligning model behavior with human preferences. It bypasses explicit reward models by directly optimizing preferences from pairwise comparisons.

Traditional methods such as reinforcement learning with human feedback (RLHF)~\cite{RLHF_1} and AI feedback (RLAIF)~\cite{RLAIF} rely on training a scalar reward model $r_\psi(x, q, y)$ from preference pairs. This reward model is typically trained using the Bradley-Terry formulation~\cite{bradley1952rank}:
\begin{equation}
\begin{aligned}
    P\left(y_w \succ y_r \mid x, q\right)
    &= \frac{\exp(r_\psi(x, q, y_w))}{\exp(r_\psi(x, q, y_w)) + \exp(r_\psi(x, q, y_r))} \\
    &= \sigma\left( r_\psi(x, q, y_w) - r_\psi(x, q, y_r) \right),
\end{aligned}
\end{equation}
where $(x, q, y_w, y_r)$ is sampled from the preference data distribution $\mathcal{D}$, and $\sigma(z) = \frac{1}{1 + \exp(-z)}$ denotes the sigmoid function. $y_w$ and $y_r$ denote the preferred and dispreferred responses, respectively. The reward model $r_\psi(x, q, y)$ is then trained to maximize the log-likelihood of correctly ranking the preferred response over the dispreferred one during optimization:
\begin{equation}
    \underset{r_\psi}{\min} \,
    \mathbb{E}_{(x, q, y_w, y_r) \sim \mathcal{D}}
    \left[
      - \log \sigma \left(
        r_\psi(x, q, y_w) - r_\psi(x, q, y_r)
      \right)
    \right].
\end{equation}
After training, the learned reward model $r_\psi(x, q, y)$ is used to guide the fine-tuning of the policy $\pi_\theta$. Specifically, the policy is optimized to generate high-reward responses while minimizing divergence from a fixed reference policy $\pi_\text{ref}$, typically using KL-regularized objectives:

\begin{equation}
\textstyle
\underset{\pi_\theta}{\min} \;
 \mathbb{E}_{(x, q) \sim \mathcal{D},\, y^* \sim \pi_\theta} \big[
   - \big( r_\psi(x, q, y^*)  -
    \alpha \cdot \mathbb{D}_{\mathrm{KL}}(
      \pi_\theta(y^* \mid x, q)
      \|
      \pi_{\text{ref}}(y^* \mid x, q))
   \big)
\big].
\end{equation}

where $\alpha$ controls the strength of KL regularization, which ensures alignment with the learned preferences. Rather than relying on the explicitly trained reward model, DPO~\cite{rafailov2023direct} simplifies the learning process by leveraging the insight that the optimal policy can be expressed in closed form using relative log-likelihoods under $\pi_\theta$ and $\pi_\text{ref}$:
\begin{equation}
\label{eq:traditional_DPO}
\textstyle
\underset{\pi_\theta}{\min} \;
 \mathbb{E}_{(x, q, y_w, y_r) \sim \mathcal{D}}
 \big[
    - \log \sigma \big(
      \alpha \log
      \frac{\pi_\theta(y_w \mid x, q)}{\pi_{\text{ref}}(y_w \mid x, q)}
      -
      \alpha \log
      \frac{\pi_\theta(y_r \mid x, q)}{\pi_{\text{ref}}(y_r \mid x, q)}
    \big)
\big].
\end{equation}
This formulation enables direct policy optimization from preference pairs, aligning the output probabilities with human preferences and improving alignment stability and sample efficiency compared to RL-based approaches.

\section{Method}
\label{sec:method}
We propose a token-adaptive min-max strategy with perturbations on visual-agnostic tokens and a frequency-based regularizer for improved alignment. An overview is shown in \cref{fig:method}, and the detailed algorithm is provided in the Appendix.\footnote{Full pseudocode is given in Appendix, Section~8.}
We first define notation used throughout: $x$ denotes the visual input, $q = \{q_1, \dots, q_m\}$ the tokenized textual prompt, $y_w$ and $y_r$ the preferred and dispreferred responses, $\pi_\theta$ the trainable policy, $\pi_{\text{ref}}$ the frozen reference policy, and $\mathcal{D}$ the preference dataset.

\begin{figure}[tb]
        \begin{center}
        \includegraphics[width=\linewidth]{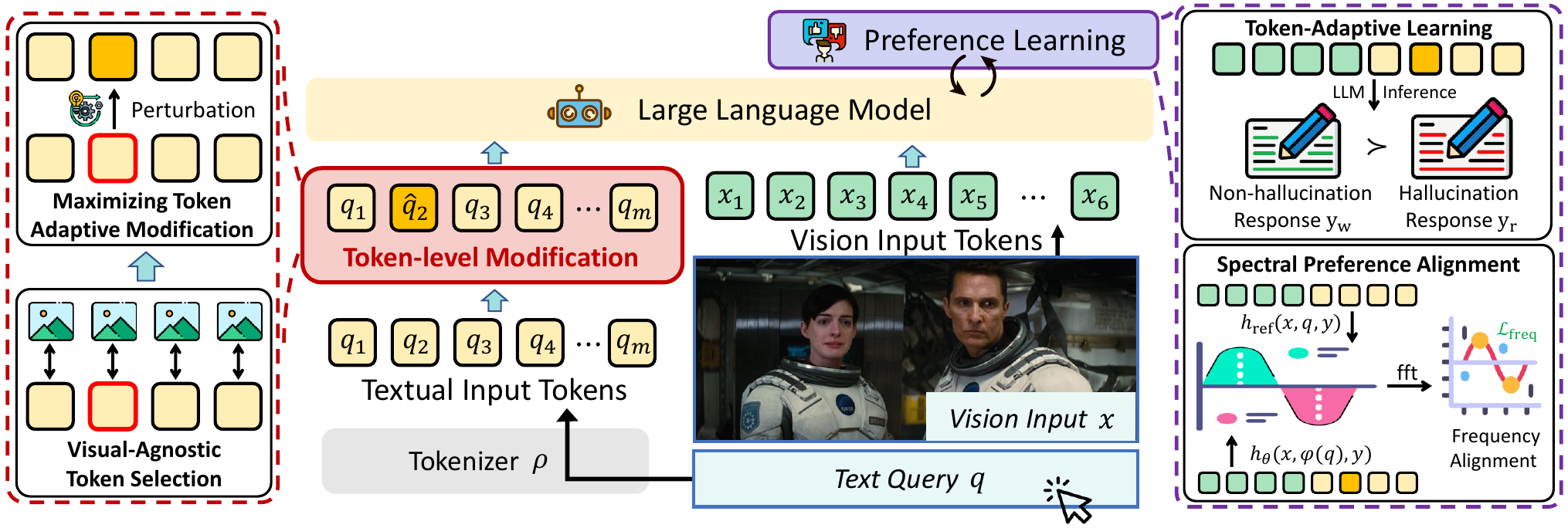}
        \end{center}
        \vspace{-1.5em}
        \caption{
        Overview of \textbf{TARS}. TARS reformulates preference optimization as a Min--Max problem:
        (1) The maximization branch perturbs visual-agnostic tokens to simulate semantically shifted contexts (\textcolor{DeepRed}{red dashed box});
        (2) The minimization branch fine-tunes the model to align with human preferences via the DPO objective (\textcolor{DeepPurple}{purple dashed box}).
        TARS encourages the model to attend to causally grounded visual signals rather than spurious correlations, thereby reducing hallucinations.
        }
        \label{fig:method}
        \vspace{-0.5em}
\end{figure}

\subsection{Min-Max Reformulation of DPO}
\label{sec:min-max-reformulation}
To address the limitations of traditional DPO, we reformulate preference optimization as a \textit{token-adaptive min-max game}.
The inner maximization introduces a controlled token-level perturbation function \(\varphi(\cdot)\), which modifies selected tokens in the prompt $q$ to induce input distribution shifts. The outer minimization then aligns the policy \(\pi_\theta\) with preference signals under these perturbations.
Formally, we define the min--max preference objective as:
\begin{equation}
\label{eq:minmax_dpo}
\min_{\pi_\theta}
\max_{\varphi \in \Phi(\mathcal{A})}\;
\mathbb{E}_{(x,q,y_w,y_r)\sim\mathcal{D}}
\left[
\mathcal{L}_{\text{TARS}}\bigl(x,\varphi(q),y_w,y_r\bigr)
\right],
\end{equation}
where \(\varphi\) is constrained to modify only visually agnostic tokens. Specifically, \(\Phi(\mathcal{A})\) denotes the set of admissible perturbation functions:
\(\Phi(\mathcal{A}) = \{\varphi \mid \{i \mid \varphi(q_i) \neq q_i\} \subseteq \mathcal{A}(x,q)\}\),
where \(\mathcal{A}(x,q)\) is the set of token indices identified as visually agnostic (defined in \cref{eq:perturbation}).
This min--max objective promotes preference alignment under distributional shifts, helping to mitigate spurious correlations and reduce hallucinated outputs.

\subsection{Maximizing with Token Perturbations}
\label{sec:max_token}
As shown in~\cref{eq:traditional_DPO}, DPO aligns models with preferred responses via log-likelihood ratios against a reference model. However, we observe that this formulation can encourage overfitting to superficial patterns such as frequent phrases and stylistic tokens, which in turn reduce effective alignment with the visual context and hinder robust multimodal understanding~\cite{setlur2024rl, CHIP}.

To counter this, we apply token-wise maximization to introduce distribution shifts and reduce overfitting to preference signals. Formally, we define:
\begin{equation}
    \label{eq:perturbation_obj}
    \varphi(q) = \underset{\varphi \in \Phi(\mathcal{A})}{\arg\max} \; \mathrm{Sim}(\varphi(q), q),
\end{equation}
where \(\Phi(\mathcal{A})\) denotes allowable perturbations constrained to \(\mathcal{A}(x,q)\), and \(\mathrm{Sim}(\varphi(q), q)\) measures token-level deviation. In practice, we approximate \(\varphi^*(q)\) by applying token-level transformations:
\begin{equation}
    \label{eq:perturb_method}
    \varphi(q) = \left\{ \mathbb{I}[i \in \mathcal{A}(x, q)] \cdot \varphi(q_i) + \mathbb{I}[i \notin \mathcal{A}(x, q)] \cdot q_i \right\}_{i=1}^{|q|},
\end{equation}
where \(\varphi(q_i)\) denotes the perturbed token, constructed using either masking (replacing with \texttt{[MASK]}) or synonym substitution (implementation details are in Appendix, Section~7). \(\mathbb{I}[\cdot]\) is the indicator function. This approximation simulates worst-case alignment uncertainty while preserving semantic integrity.

To preserve semantics, we restrict changes to visual-agnostic tokens with minimal impact on cross-modal alignment. We compute token-level visual relevance using a visual encoder $\mathcal{G}_v(\cdot)$ and a text encoder $\mathcal{G}_t(\cdot)$ (both instantiated by the CLIP encoder~\cite{radford2021learning}) as the cosine similarity between visual features $\mathcal{G}_v(x) \in \mathbb{R}^{d}$ and each token embedding $\mathcal{G}_t(q_i) \in \mathbb{R}^{d}$.
We then identify a set $\mathcal{A}$ of $N_t$ visually agnostic tokens with the lowest cross-modal alignment scores:
\begin{equation}
\label{eq:perturbation}
\textstyle
\mathcal{A} = \text{Top}_{N_t}\!\big(-\, \mathcal{G}_v(x) \mathcal{G}_t(q_i)^{\top}\big), \quad
N_t = \lfloor \omega \cdot \Delta P^{-1} \rfloor + 1.
\end{equation}
where \(\lfloor \cdot \rfloor\) denotes the floor operation and \(\omega\) is a scaling coefficient controlling perturbation intensity. The matrix $P \in \mathbb{R}^{m}$ is the negated similarity score vector, with $P_i = -\mathcal{G}_v(x) \cdot \mathcal{G}_t(q_i)^T$. The confidence margin $\Delta P = \max_j P_j - \max_{k \neq j} P_k$ quantifies the predictive uncertainty of the cross-modal alignment: confident predictions (large $\Delta P$) lead to fewer perturbations, while greater uncertainty (small $\Delta P$) induces broader variation. An empirical comparison of this adaptive selection strategy against random and uniform perturbation baselines is provided in the Appendix, Section~9.5.

\subsection{Spectral Regularization for Token Alignment}
\label{sec:spectral}
Token-level perturbation introduces distribution shifts, yet the supervision from preference pairs $(y_w, y_r)$ is static. This discrepancy may cause the model to learn distribution-specific artifacts under strong alignment constraints~\cite{CHIP, chowdhuryprovably}.

We align in the frequency domain rather than in the spatial domain (\eg, $\ell_2$). The key motivation is that a spatial loss $\|z - z'\|_2$ penalizes every positional deviation equally; when specific tokens are perturbed, this rigid constraint reintroduces spurious positional correlations~\cite{zhou2024explore, tianblack}. In contrast, by the shift property of the Discrete Fourier Transform, perturbing a single token $z_l$ affects every frequency bin $k$ by an additive term proportional to $e^{-2\pi i k l / L}$, whose magnitude is bounded regardless of $l$. The spectral representation thus absorbs local perturbations into a smooth global envelope, preserving dominant low-frequency semantics while tolerating position-specific noise~\cite{cooley1965algorithm}.\footnote{A formal theoretical analysis of spectral alignment under token-level adversarial perturbation, including proofs of energy dispersion and bounded semantic shift via low-frequency masking, is provided in the Appendix, Section~9.3.}

Concretely, we extract hidden states for \((x, \varphi(q), y_w)\) and contrast them with \((x, q, y_w)\) and \((x, q, y_r)\). Let \(z \in \mathbb{R}^{L \times D}\) denote a hidden-state sequence. The spectral representation is:
\begin{equation}
\footnotesize
\label{eq:fft}
    \mathcal{F}(z) = \left| \text{Re} \left[ \sum_{l=0}^{L-1} z_l \cdot e^{-2\pi i k l / L} \right] \right|_2, \ \text{for } k = 0, \dots, L{-}1,
\end{equation}
where the FFT is applied along the token axis, $\text{Re}[\cdot]$ extracts the real part, and $|\cdot|_2$ yields a scalar spectral summary. The spectral preference loss is:
\begin{equation}
\label{eq:freq_dpo}
\textstyle
\mathcal{L}_{\text{freq}}
= - \log \sigma \Big(
 \beta \big[
   \log \frac{\mathcal{F}(h_{\theta}(x, \varphi(q), y_w))}
             {\mathcal{F}(h_{\text{ref}}(x, q, y_w))}
 - \log \frac{\mathcal{F}(h_{\theta}(x, \varphi(q), y_r))}
             {\mathcal{F}(h_{\text{ref}}(x, q, y_r))}
   \big]
\Big).
\end{equation}
Here $h_{\theta}(\cdot)$ and $h_{\text{ref}}(\cdot)$ are hidden states from the policy and reference models, and $\beta$ is a scaling temperature. This objective extends DPO alignment to the spectral domain, improving frequency-aware consistency and reducing hallucinations from overfitting to fixed preferences.

\subsection{Minimization Objective in TARS}
\label{sec:joint}
We integrate the standard DPO loss with spectral regularization to yield the final TARS training objective. Given a perturbed input $\varphi(q)$ obtained from the inner maximization, and its original counterpart $q$, the overall loss is defined as:
\vspace{-1em}
\begin{equation}
\label{eq:total_loss}
\footnotesize
\mathcal{L}_{\text{TARS}}(x, q, \varphi(q), y_w, y_r)
= \;  \mathcal{L}_{\text{DPO}}(x, \varphi(q), y_w, y_r)
 + \lambda \cdot \mathcal{L}_{\text{freq}}(x, q, \varphi(q), y_w, y_r).
\end{equation}
where $\lambda$ is a weighting coefficient that balances preference alignment and spectral consistency. This joint formulation explicitly encourages the model to preserve causal alignment with preference signals under adversarial perturbation, thereby mitigating spurious correlation. Extended ablation studies on the sensitivity of $\omega$ and $\lambda$ are presented in Appendix, Section~9.

\section{Experiments}
\label{sec:ex}
\subsection{Experiment Details}
\label{sec:ex_detail}

\noindent\textbf{Experiment Setups.}
We evaluate our approach on the multimodal LLM LLaVA-v1.5~\cite{llava_origin} at both 7B and 13B scales, and on Muffin-13B~\cite{yu2023reformulating, CHIP} (Appendix Section 11). All methods are performed with greedy decoding and a temperature of 0.

For fair comparison, we carefully align our training configuration with the most data-efficient preference optimization baselines. Specifically, we randomly sample 4.8k instances from the RLHF-V-Dataset~\cite{yu2024rlhf}, consistent with OPA-DPO~\cite{OPA_DPO}, and adopt the same training strategy as CHiP-DPO~\cite{CHIP}.
All models are trained on eight NVIDIA A100 (80GB) GPUs with identical hyperparameters. We set \(\alpha=1\) in~\cref{eq:traditional_DPO} and \(\beta=1\) in~\cref{eq:freq_dpo} for preference optimization. We implement \(\varphi(\cdot)\) using both token masking (Mask) and replacement (Replace) strategies in~\cref{eq:perturb_method}, and set the perturbation constraint strength to \(\omega = 0.1\) in the adversarial min-max formulation~\cref{eq:perturbation}. We use a frequency-domain loss weight of \(\lambda = 0.1\) in~\cref{eq:total_loss}. Full implementation details and ablation studies are reported in the Appendix.\footnote{See Appendix, Sections~7 and~9 for configurations and hyperparameter ablations.}

\noindent\textbf{Evaluation Benchmarks.}
We evaluate TARS across four established hallucination benchmarks spanning generative and discriminative settings: AMBER~\cite{wang2023amber} for fine-grained generative hallucination (CHAIR~\cite{rohrbach2018object}, Cover, Hal-Rate, Cog), MMHal~\cite{sun2023aligning} for VQA hallucination scored by GPT-4V, OBJHal~\cite{yu2024rlhf} for captioning hallucination (response-level CR$_s$ and object-level CR$_i$), and POPE~\cite{li2023evaluating} for binary discriminative object hallucination. Detailed benchmark descriptions and evaluation protocols are provided in Appendix, Section~7.

\noindent\textbf{Baseline Methods.}
We compare against two categories:

\noindent\textbf{(1) Advanced multimodal foundation models:} Intern-VL2.5-7B~\cite{chen2024expanding}, Qwen-VL2.5-8B~\cite{bai2025qwen2}, DeepSeek-VL2-27B~\cite{wu2024deepseek}, and GPT-4o~\cite{hurst2024gpt}.

\noindent\textbf{(2) LLaVA-v1.5 with RL techniques:} We evaluate multiple RL-based approaches applied to both the 7B and 13B variants of LLaVA-v1.5, including RLHF~\cite{RLHF}, RLAIF~\cite{RLAIF}, HALVA~\cite{HALVA}, as well as three state-of-the-art methods based on direct preference optimization (DPO): DPO~\cite{DPO_MLLMs}, CHiP-DPO~\cite{CHIP}, and OPA-DPO~\cite{OPA_DPO}.
A comparison of algorithmic properties is provided in~\cref{tb:comparison_PO}.

\begin{table}[tb]
    \caption{Comparison across benchmarks. We evaluate SOTA MLLMs as references, denoted by $\S$. For algorithms with available checkpoints, re-tested results are marked with $\dagger$; for those without, we reproduce results using settings from \cite{CHIP, li2024multi}, denoted by $\ddagger$. \textbf{Bold} denotes the best performance, and \underline{underlined} denotes the second-best.}
    \vspace{-1em}

    \label{tb:sota}
    \resizebox{\linewidth}{!}
    {
        \begin{tabular}{lcccccccccc}
        \toprule[1.2pt]
        \multirow{2}{*}{\textbf{Algorithm}} &
        \multicolumn{4}{c}{\textbf{AMBER}} &
        \multicolumn{2}{c}{\textbf{MMHal}} &
        \multicolumn{2}{c}{\textbf{POPE}} &
        \multicolumn{2}{c}{\textbf{OBJHal}} \\
        \cmidrule[0.5pt](lr){2-5} \cmidrule[0.5pt](lr){6-7} \cmidrule[0.5pt](lr){8-9} \cmidrule[0.5pt](lr){10-11}
        & \textbf{CHAIR$\downarrow$} & \textbf{Cover$\uparrow$} & \textbf{Hal-Rate$\downarrow$} & \textbf{Cog$\downarrow$} & \textbf{Score$\uparrow$} & \textbf{Hal-Rate$\downarrow$} & \textbf{Acc$\uparrow$} & \textbf{Pre$\uparrow$} & \textbf{CR$_\text{s}$$\downarrow$} & \textbf{CR$_\text{i}$$\downarrow$} \\
        \midrule
        Intern-VL2.5-7B~\cite{chen2024expanding}$^\S$
        &7.9&54.7&37.1&3.2&3.54&0.26&-&-&36.0&9.1 \\
        Qwen-VL2.5-8B~\cite{bai2025qwen2}$^\S$
        &4.6&54.6&21.1&1.3&3.29&0.27&-&-&40.7&8.6 \\
        DeepSeek-VL2-27B~\cite{wu2024deepseek}$^\S$
        &2.4&56.6&16.3&0.9&2.84&0.27&-&-&10.0&7.0 \\
        GPT-4o~\cite{hurst2024gpt}$^\S$
        &2.5&60.9&17.6&0.8&3.87&0.24&-&-&29.3&6.7 \\
        \midrule[0.4pt]
        \textbf{LLaVA-v1.5-7B}~\cite{llava_origin}$^\S$
        &7.6&51.7&35.4&4.2&2.02&0.61&80.0&61.8&54.0&15.8 \\
        \hspace{3pt} + \textit{RLHF}~\cite{RLHF}$^\dagger$
        &8.3&52.2&41.8&4.5&1.93&0.67&82.0&69.3&56.0&15.2 \\
        \hspace{3pt} + \textit{RLAIF}~\cite{RLAIF}$^\dagger$
        &3.0&50.3&16.5&1.0&\textbf{2.89}&\textbf{0.42}&\underline{88.1}&88.0&13.7&4.2 \\
        \hspace{3pt} + \textit{HALVA}~\cite{HALVA}$^\dagger$
        &6.9&52.8&33.2&3.5&2.12&0.59&87.5&79.6&47.3&14.6 \\
        \hspace{3pt} + \textit{DPO}~\cite{li2024multi}$^\ddagger$
        &4.9&56.6&26.4&2.5&2.19&0.61&87.8&82.0&14.0&5.0 \\
        \hspace{3pt} + \textit{CHiP-DPO}~\cite{CHIP}$^\ddagger$
        &2.9&{57.3}&19.9&1.0&2.32&0.57&81.1&91.8&\textbf{7.3}&4.3 \\
        \hspace{3pt} + \textit{OPA-DPO}~\cite{OPA_DPO}$^\dagger$ &2.7&47.4&\textbf{12.5}&{0.9}&\underline{2.78}&0.46&87.4&86.2&13.3&4.5 \\
        \rowcolor{orange!15}  \hspace{3pt} + TARS~(Mask)&\underline{2.4}&\textbf{59.6}&\underline{13.2}&\textbf{0.4}&2.48&\underline{0.45}&\textbf{88.7}&\textbf{97.5}&\underline{12.0}&\textbf{3.2} \\
        \rowcolor{orange!15}  \hspace{3pt} + TARS~(Replace)
        &\textbf{2.1}&\underline{59.3}&14.9&\underline{0.7}&2.54&0.46&87.9&\underline{97.0}&13.4&\underline{3.3} \\
        \midrule[0.8pt]
        \textbf{LLaVA-v1.5-13B}~\cite{llava_origin}$^\S$
        &6.7&52.1&32.5&3.5&2.39&0.53&74.6&55.2&50.0&14.5 \\
        \hspace{3pt} + \textit{RLHF}~\cite{RLHF}$^\dagger$
        &7.1&51.4&36.3&3.6&2.10&0.67&83.6&71.2&46.7&11.6 \\
        \hspace{3pt} + \textit{HALVA}~\cite{HALVA}$^\dagger$
        &6.5&53.4&30.1&3.3&2.28&0.56&86.8&75.6&42.7&12.1 \\
        \hspace{3pt} + \textit{DPO}~\cite{li2024multi}$^\ddagger$
        &4.1&56.7&24.3&2.2&2.48&0.50&85.2&84.3&19.0&7.2 \\
        \hspace{3pt} + \textit{CHiP-DPO}~\cite{CHIP}$^\ddagger$
        &3.8&{58.6}&20.8&1.7&2.70&0.46&86.6&74.9&30.0&6.2 \\
        \hspace{3pt} + \textit{OPA-DPO}~\cite{OPA_DPO}$^\dagger$
        &2.8&48.4&\underline{13.5}&{1.0}&\textbf{3.02}&\textbf{0.40}&\underline{87.2}&80.7&18.3&5.1 \\
        \rowcolor{orange!15}  \hspace{3pt} + TARS~(Mask)
        &\textbf{2.1}&\textbf{59.8}&\textbf{12.5}&\textbf{0.6}&\underline{2.89}&\underline{0.45}&\textbf{87.6}&\textbf{93.0}&\textbf{14.6}&\textbf{2.8} \\
        \rowcolor{orange!15}  \hspace{3pt} + TARS~(Replace)
        &\textbf{2.1}&\underline{59.4}&13.6&\underline{0.7}&2.63&0.47&86.9&\underline{92.5}&\underline{14.9}&\underline{3.4} \\
        \bottomrule[1.2pt]
        \end{tabular}
    }
        \vspace{-1.5em}
\end{table}

\subsection{Evaluation on Hallucination Benchmarks}
\cref{tb:sota} presents results across four hallucination benchmarks. We adopt token masking and synonym replacement as perturbation strategies (extended results on Muffin-13B are in Appendix, Section~11).

\noindent\textbf{(1) Consistent hallucination mitigation across benchmarks.}
On the 7B scale, TARS lowers the AMBER Hal-Rate from 35.4\% to 13.2\% (--22.2~pp) while raising Cover from 51.7\% to 59.6\% (+7.9~pp) and reducing Cog from 4.2 to 0.4. On OBJHal, CR$_\text{s}$ drops sharply from 54.0\% to 12.0\%.

\noindent\textbf{(2) Strong data efficiency under limited supervision.}
Using only 4.8k public preference samples without expert feedback (\cref{tb:comparison_PO}), TARS outperforms OPA-DPO on AMBER-13B in both Cover (+11.4~pp) and Hal-Rate (--1.0~pp).

\noindent\textbf{(3) Scalability across model sizes.}
From 7B to 13B, CHAIR improves from 2.4 to 2.1, Hal-Rate drops from 13.2\% to 12.5\%, and Cog stays below 0.7. TARS-13B consistently surpasses all 13B baselines by 1.0--1.5~pp in Hal-Rate.

\noindent\textbf{(4) Competitiveness with larger proprietary models.}
At 13B, TARS approaches GPT-4o in Cover (59.8\% vs.\ 60.9\%) while achieving a notably lower Hal-Rate (12.5\% vs.\ 17.6\%), and remains competitive with DeepSeek-VL2-27B in both metrics despite its smaller scale. POPE accuracy reaches 88.7\% (+8.7~pp over LLaVA-7B), confirming preserved factual grounding (breakdowns in Appendix, Section~10).\footnote{TARS also improves fine-grained discriminative detection (Appendix, Section~10.1) and general multimodal understanding (Appendix, Section~10.2).}

\begin{figure}[tb]
    \centering
    \includegraphics[width=\linewidth]{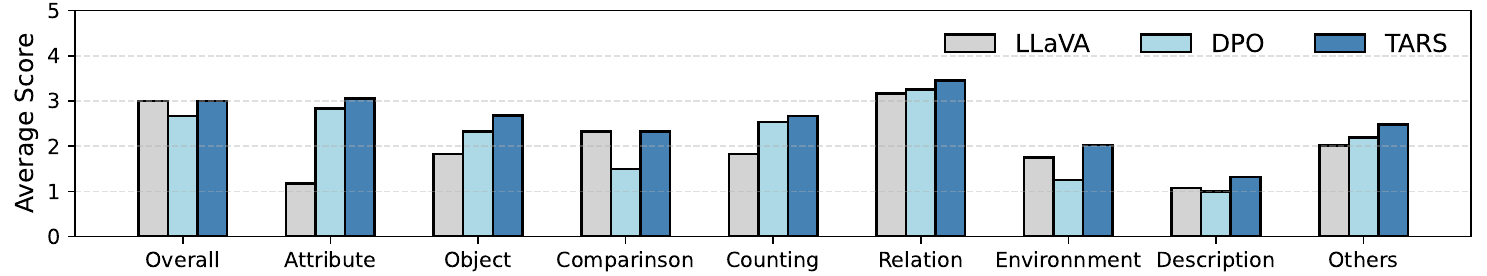}
    \vspace{-2em}
    \caption{Comparison of average scores across question categories on MMHal. TARS achieves consistently higher scores, demonstrating stronger visual grounding.}
    \label{fig:MMH_score}
    \vspace{-0em}
\end{figure}

\begin{figure}[tb]
    \centering
    \begin{minipage}[t]{0.53\linewidth}
        \centering
        \vspace{-2em}
        \captionof{table}{Ablation of token-level perturbation (TP), cross-modal alignment score (CAS), and spectral preference alignment (SPA).}
        \label{tb:ablation}
        \resizebox{\linewidth}{!}{
        \begin{tabular}{lccccc}
            \toprule[1.0pt]
            \multirow{2}{*}{\textbf{Algorithm}} &
            \multicolumn{3}{c}{\textbf{AMBER}} &
            \multicolumn{2}{c}{\textbf{OBJHal}} \\
            \cmidrule[0.5pt](lr){2-4} \cmidrule[0.5pt](lr){5-6}
            & \textbf{Cover$\uparrow$} & \textbf{Hal-Rate$\downarrow$} & \textbf{Cog$\downarrow$} &  \textbf{CR$_\text{s}$$\downarrow$} & \textbf{CR$_\text{i}$$\downarrow$} \\
            \midrule
            \textbf{LLaVA-v1.5-7B}~\cite{llava_origin}&51.7&35.4&4.2&54.0&15.8\\
            \midrule
            \rowcolor{orange!15} \textbf{TARS}   &\textbf{59.6}&\textbf{13.2}&\textbf{0.4}&\textbf{12.0}&\textbf{3.2} \\
            w/o \textbf{TP} &56.6&26.4&2.5&14.0&5.0\\
            w/o \textbf{CAS}&55.9&17.7&1.3&12.7&3.5\\
            w/o \textbf{SPA}&58.3&15.1&0.7&12.5&3.7 \\
            w/o \textbf{CAS}\textbf{\&}\textbf{SPA} &55.1&18.5&1.5&12.6&3.8 \\
            \bottomrule[1.0pt]
        \end{tabular}
        }
    \end{minipage}
    \hfill
    \begin{minipage}[t]{0.46\linewidth}
        \centering
        \vspace{-0.5em}
        \includegraphics[width=\linewidth]{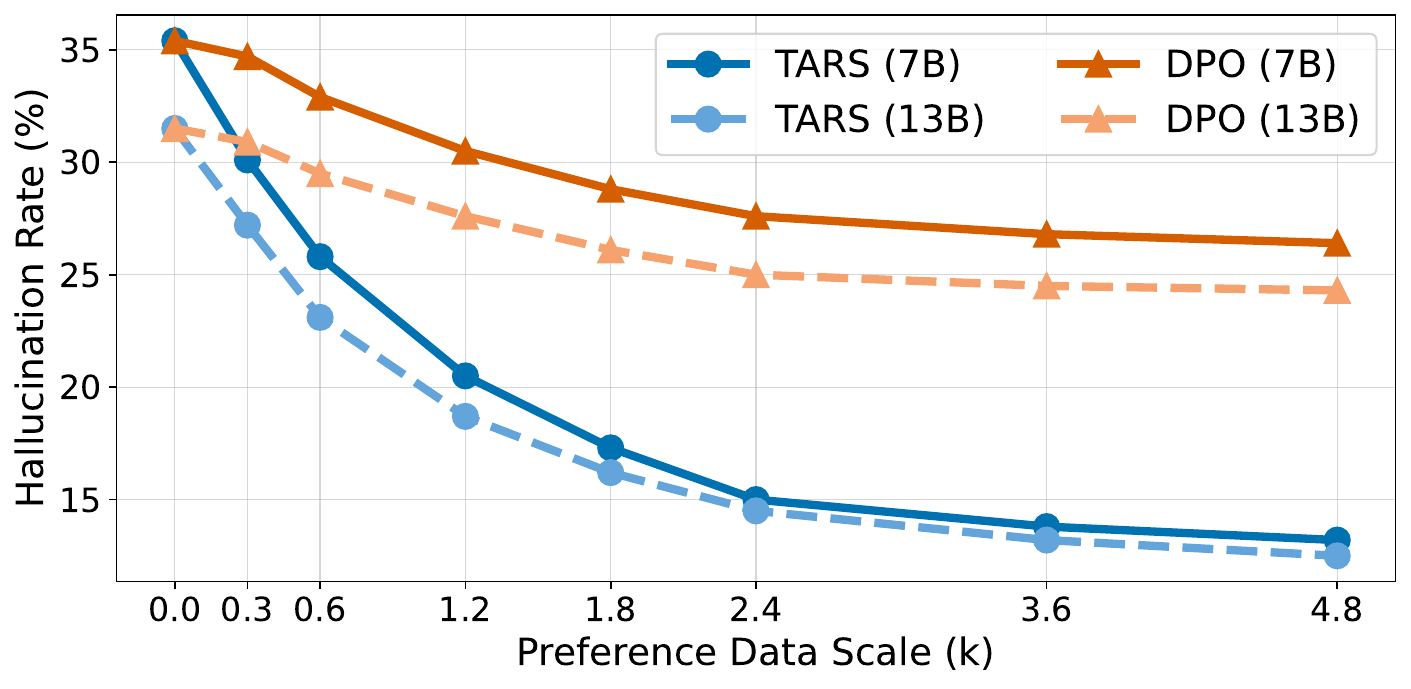}
        \vspace{-0.9em}
        \captionof{figure}{Comparison of AMBER hallucination rate versus preference data scale.}
        \label{fig:ablation-plot}
    \end{minipage}
    \vspace{-1em}
\end{figure}

\begin{figure}[tb]
    \centering
    \setlength{\tabcolsep}{1pt}
    \renewcommand{\arraystretch}{1}
    \begin{tabular}{@{}ccc@{}}
        \includegraphics[width=0.33\linewidth]{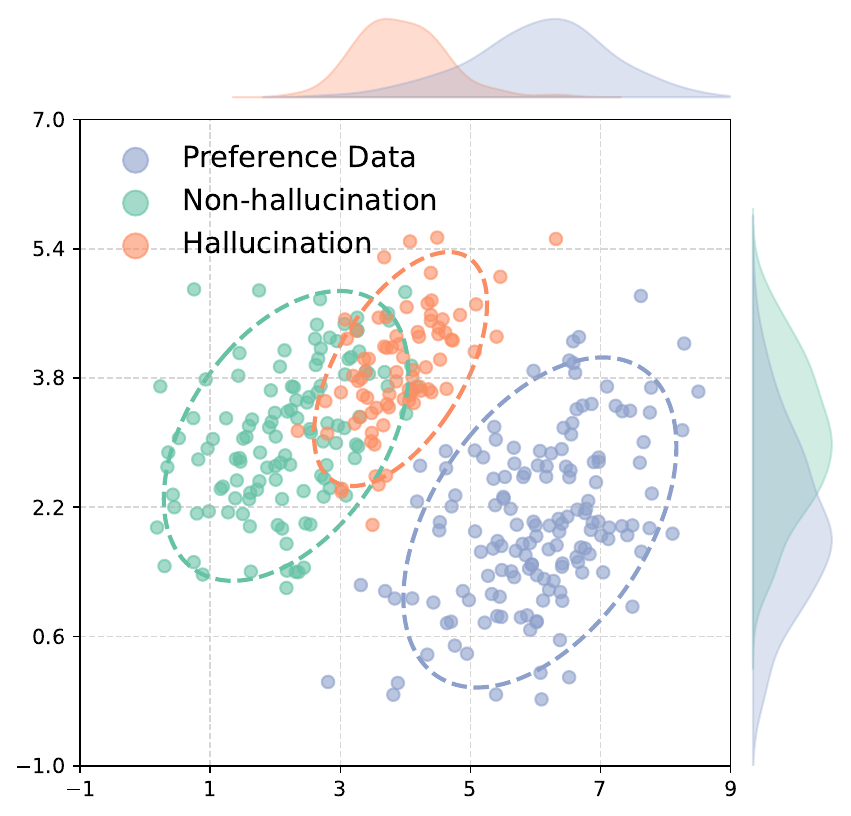} &
        \includegraphics[width=0.33\linewidth]{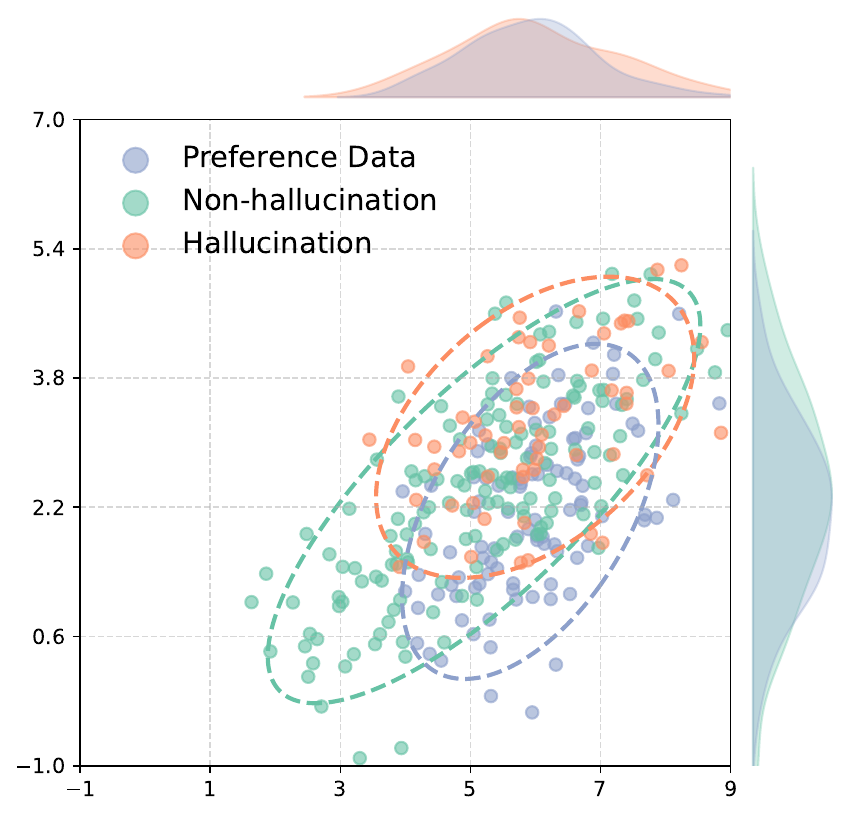} &
        \includegraphics[width=0.33\linewidth]{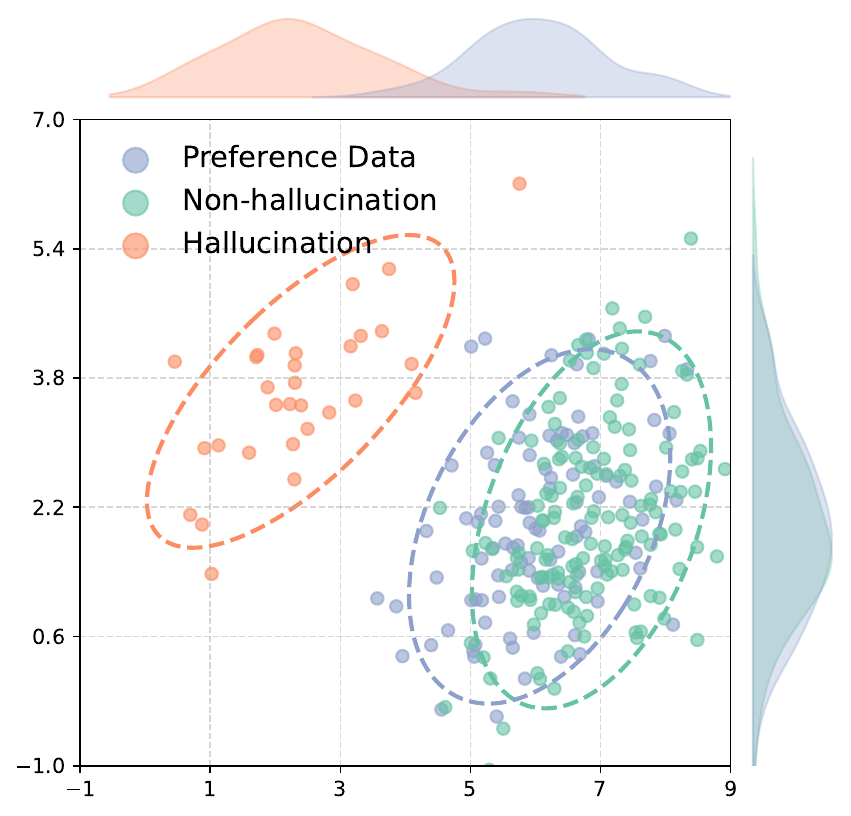} \\
        \footnotesize{(a) LLaVA} &
        \footnotesize{(b) DPO} &
        \footnotesize{(c) TARS}
    \end{tabular}
    \vspace{-0.5em}
    \caption{
    Distribution of hidden representations across preference-aligned, non-hallucinated, and hallucinated responses of different MLLMs. Top and right margins show marginal distributions along key feature dimensions. We extract representations from 100 preference training instances and 200 AMBER inputs across text and vision modalities. Responses to AMBER inputs are categorized as non-hallucinated or hallucinated based on factual coherence. TARS aligns with preference data while avoiding overfitting to spurious correlations, demonstrating superior factual fidelity.
    }
    \label{fig:feature_dist}
    \vspace{-1em}
\end{figure}

\subsection{Ablation Analyses on Component}
We analyze the key TARS components through ablations in~\cref{tb:ablation}, focusing on three elements:

\noindent\textbf{(1) Token-level perturbation (TP)} in \cref{eq:minmax_dpo}, which introduces distributional shifts and proves essential for revealing token-level vulnerabilities and improving robustness. Removing it substantially increases Cog from 0.4 to 2.5, highlighting its critical role in capturing fine-grained alignment uncertainty.

\noindent\textbf{(2) Cross-modal alignment score (CAS)} in \cref{eq:perturbation}, which targets visually agnostic tokens to preserve semantic fidelity. Its absence leads to a 4.5-point increase in hallucination and a 0.9 rise in Cog, indicating weaker suppression of spurious correlations and reduced visual grounding. A direct comparison between adaptive and random/uniform token selection is provided in the Appendix, Section~9.5, confirming the necessity of the proposed mechanism.

\noindent\textbf{(3) Spectral preference alignment (SPA)} in \cref{eq:freq_dpo}, which regularizes frequency-aware consistency across token representations. Removing it increases the hallucination rate by 1.9 points and CR$_\text{i}$ from 3.2 to 3.7, suggesting degraded fine-grained factual grounding and less stable preference alignment. A comparison of spectral alignment against spatial alternatives ($\ell_2$, cosine) is reported in the Appendix, Section~9.4.

\subsection{Ablation Analyses on Preference Scale Impact}
We examine how preference data scale influences alignment efficiency (\cref{fig:ablation-plot}). TARS consistently outperforms DPO across all scales and exhibits sharper gains in the low-data regime: from 0 to 1.8k examples, the 7B and 13B variants reduce hallucination rates by over 15 pp, indicating that TARS captures the core alignment signal early. Beyond 3.6k examples, marginal gains saturate while performance remains stable, highlighting strong data efficiency for practical scenarios where preference annotations are costly or scarce.
This behavior is consistent with the dual-space regularization discussed in \cref{sec:joint}: by enforcing alignment in both the output probability space and the hidden representation space, TARS extracts richer supervision from each preference pair, reducing the sample complexity needed to reach competitive performance. In contrast, standard DPO requires substantially more data to compensate for its tendency to overfit to surface-level patterns, as reflected by the persistent hallucination rate gap across all data scales in \cref{fig:ablation-plot}.

\subsection{Stability of Semantic Representations}
We analyze how preference optimization reshapes hidden-state distributions in \cref{fig:feature_dist}. TARS yields a more structured latent space in which hallucinated and preference-aligned representations are clearly separated, whereas DPO produces entangled clusters that interleave hallucinated and preference features, indicating overfitting to superficial signals. Crucially, TARS selectively aligns non-hallucinated responses with preference features while isolating hallucinated content, creating a semantically faithful representation space that reinforces only factually grounded outputs.

The marginal distributions (top and right panels in \cref{fig:feature_dist}) further reveal the mechanism behind this separation. Under DPO, hallucinated and preference-aligned marginals overlap heavily, reflecting a representation space where surface-level patterns dominate over grounded semantics. In contrast, TARS produces near-disjoint marginals for hallucinated content while maintaining tight overlap between non-hallucinated and preference-aligned distributions. This selective clustering indicates that spectral regularization combined with adversarial perturbation shapes a representation geometry that naturally distinguishes factual from hallucinated outputs, without requiring explicit labels during training.

\subsection{Training Dynamics Analysis}
\label{sec:training_dynamics}
\begin{figure}[tb]
    \centering
    \includegraphics[width=\linewidth]{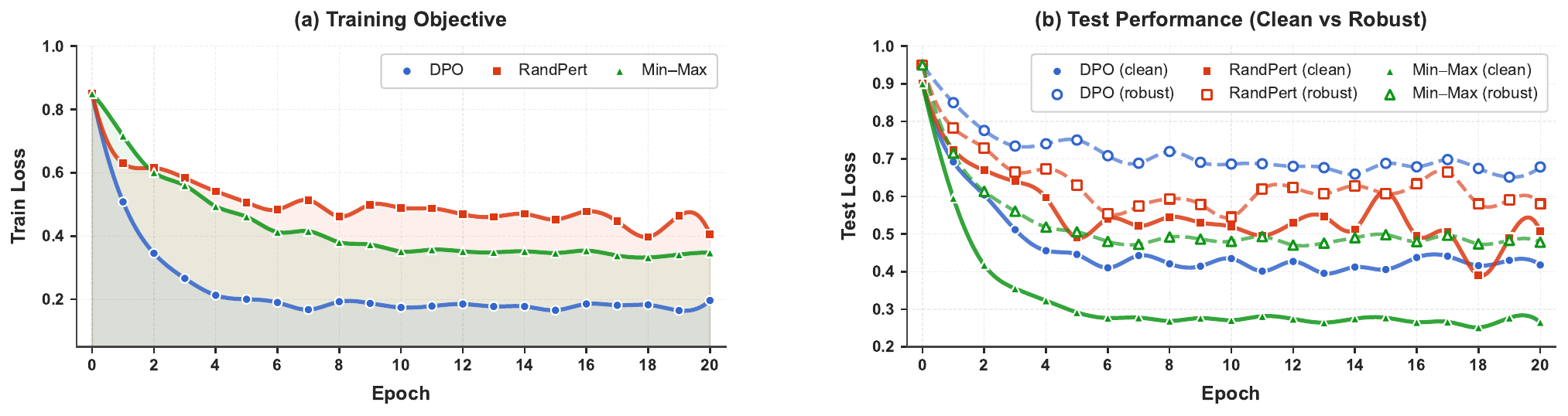}
    \vspace{-2em}
    \caption{Training dynamics of DPO, RandPert, and TARS (Min-Max); all losses are normalized to $[0,1]$. \textbf{(a)} Training objective: DPO converges fastest but overfits; Min-Max maintains a higher loss reflecting adversarial difficulty. \textbf{(b)} Test performance on clean and robust (perturbed) samples: Min-Max achieves the lowest test loss on both splits, while DPO exhibits a large clean--robust gap indicating poor robustness.}
    \label{fig:minmax_demo}
    \vspace{-1em}
\end{figure}

To validate the min--max formulation (\cref{eq:minmax_dpo}), we compare training dynamics of standard DPO, random perturbation (RandPert), and TARS (Min-Max) on LLaVA-v1.5-7B over 20 epochs (\cref{fig:minmax_demo}; all losses normalized to $[0,1]$).
DPO converges to the lowest training loss ($\sim$0.2) but exhibits the worst robust-sample test loss (0.65--0.75), confirming severe overfitting to static preference patterns~\cite{DPO_MLLMs}.
RandPert shows intermediate test performance but with heavy oscillation, as indiscriminate perturbation yields inconsistent gradients.
TARS maintains a moderately higher training loss (0.35--0.4) yet achieves the \emph{lowest} test loss on both clean ($<$0.3) and robust ($<$0.5) splits with stable convergence. This confirms that the min--max formulation acts as an implicit regularizer: adversarial token-level shifts during training prevent premature convergence and yield alignment features that generalize to unseen and perturbed data.

\subsection{Comparison with Data Augmentation}
\label{sec:augmentation_compare}
We compare TARS against data augmentation baselines across all four benchmarks (\cref{tb:augmentation}).\footnote{See Appendix, Section~10.3 for detailed augmentation protocols and analysis.} We consider paraphrasing via back-translation and LLM-based instruction augmentation at $1\times$ and $5\times$ expansion rates, all applied on top of standard DPO training.
As shown in \cref{tb:augmentation}, augmentation alone improves coverage and moderately reduces hallucination, yet even $5\times$ LLM augmentation still falls short of TARS in Hal-Rate (16.0\% vs.\ 13.2\%) and Cog (1.2 vs.\ 0.4). The trend is consistent across MMHal and OBJHal: augmentation narrows the gap but cannot close it without TARS's token-level robustness mechanism. Combining TARS with augmentation yields the best results on every metric (CHAIR~2.1, Hal-Rate~12.3\%, CR$_\text{s}$~10.6\%), confirming that the two strategies target complementary error sources.

\noindent\textbf{Discussion.}
Across all experiments, TARS's improvements manifest systematically across generative (AMBER, OBJHal), discriminative (POPE), and open-ended (MMHal) evaluations. The ablation studies (\cref{tb:ablation}) confirm that each component contributes independently, the scale analysis (\cref{fig:ablation-plot}) demonstrates strong data efficiency, and the training dynamics (\cref{fig:minmax_demo}) validate the regularization effect of the min-max formulation. This convergence of evidence across complementary evaluation axes supports the claim that token-adaptive perturbation with spectral regularization addresses a fundamental limitation of static preference optimization rather than exploiting benchmark-specific biases.

\begin{table}[tb]
  \centering
  \caption{Comparison of TARS against data augmentation baselines on four hallucination benchmarks. \textit{Paraphrasing} denotes back-translation-based response diversification; \textit{LLM Aug.} denotes LLM-generated preferred responses, where $1\times$ and $5\times$ indicate the expansion factor relative to the original 4.8k preference set. All methods build on standard DPO. TARS is complementary to augmentation strategies.}
  \label{tb:augmentation}
  \vspace{-1em}
  \setlength{\tabcolsep}{3pt}
  \renewcommand{\arraystretch}{1.1}
  \resizebox{\linewidth}{!}{
  \footnotesize{
    \begin{tabular}{lcccccccccc}
        \toprule
        \multirow{2}{*}{\textbf{Method}} &
        \multicolumn{4}{c}{\textbf{AMBER}} &
        \multicolumn{2}{c}{\textbf{MMHal}} &
        \multicolumn{2}{c}{\textbf{POPE}} &
        \multicolumn{2}{c}{\textbf{OBJHal}} \\
        \cmidrule[0.5pt](lr){2-5} \cmidrule[0.5pt](lr){6-7} \cmidrule[0.5pt](lr){8-9} \cmidrule[0.5pt](lr){10-11}
        & CHAIR$\downarrow$ & Cover$\uparrow$ & Hal-Rate$\downarrow$ & Cog$\downarrow$
        & Score$\uparrow$ & Hal-Rate$\downarrow$
        & Acc$\uparrow$ & Pre$\uparrow$
        & CR$_\text{s}$$\downarrow$ & CR$_\text{i}$$\downarrow$ \\
        \midrule
        LLaVA-7B-DPO
        & 4.9 & 56.6 & 26.4 & 2.5 & 2.19 & 0.61 & 87.8 & 82.0 & 14.0 & 5.0 \\
        + Paraphrasing
        & 4.5 & 57.9 & 24.8 & 2.3 & 2.24 & 0.58 & 87.9 & 83.5 & 13.6 & 4.8 \\
        + LLM Aug.\ (1$\times$)
        & 4.1 & 58.8 & 22.9 & 2.0 & 2.30 & 0.55 & 88.1 & 85.8 & 13.2 & 4.5 \\
        + LLM Aug.\ (5$\times$)
        & 3.1 & 59.3 & 16.0 & 1.2 & 2.38 & 0.50 & 88.3 & 90.4 & 12.6 & 4.0 \\
        \rowcolor{orange!20}
        + TARS
        & {2.4} & {59.6} & {13.2} & {0.4} & {2.48} & {0.45} & {88.7} & {97.5} & {12.0} & {3.2} \\
        \rowcolor{orange!20}
        + TARS \& Para.
        & 2.4 & 60.1 & 12.8 & 0.4 & 2.50 & 0.45 & 88.9 & 97.5 & 11.6 & 3.0 \\
        \rowcolor{orange!20}
        + TARS \& Aug.\ (1$\times$)
        & 2.2 & 60.5 & 12.5 & 0.3 & 2.53 & 0.43 & 88.9 & 97.7 & 11.2 & 2.9 \\
        \rowcolor{orange!20}
        + TARS \& Aug.\ (5$\times$)
        & \textbf{2.1} & \textbf{60.8} & \textbf{12.3} & \textbf{0.2} & \textbf{2.55} & \textbf{0.42} & \textbf{89.1} & \textbf{97.8} & \textbf{10.6} & \textbf{2.7} \\
        \bottomrule
    \end{tabular}
  }}
\end{table}

\section{Related Work}
\label{sec:related}

\begin{table}[t]
    \caption{Comparison of preference optimization strategies. TARS achieves hallucination mitigation and causal alignment with minimal data and no expert feedback.}
    \label{tb:comparison_PO}
    \vspace{-1em}
    \resizebox{\linewidth}{!}
    {
        \begin{tabular}{lccccc}
        \toprule[1.2pt]
        \textbf{Algorithm} & \textbf{Data Size} & \textbf{Feedback} & \textbf{Reward-Free} & \textbf{Hal. Mitigation} & \textbf{Causal Align.} \\
        \midrule[0.4pt]
        \textbf{LLaVA-v1.5}~\cite{llava_origin}
                               & -    & -            & \xmark & \xmark & \xmark \\
        \hspace{3pt} + \textit{RLHF}~\cite{RLHF}       & 122k & self-reward  & \cmark & \xmark & \xmark \\
        \hspace{3pt} + \textit{RLAIF}~\cite{RLAIF}     & 16k  & LLaVA-Next   & \xmark & \cmark & \xmark \\
        \hspace{3pt} + \textit{HALVA}~\cite{HALVA}     & 22k  & GPT-4V       & \xmark & \xmark & \xmark \\
        \hspace{3pt} + \textit{DPO}~\cite{li2024multi} & 5k   & self-reward  & \cmark & \cmark & \xmark \\
        \hspace{3pt} + \textit{CHiP-DPO}~\cite{CHIP}   & 5k   & self-reward  & \cmark & \cmark & \xmark \\
        \hspace{3pt} + \textit{OPA-DPO}~\cite{OPA_DPO} & 4.8k & GPT-4V       & \xmark & \cmark & \xmark \\
        \hspace{3pt} + \textbf{\textit{TARS (Ours)}}   & 4.8k & self-reward  & \cmark & \cmark & \cmark \\
        \bottomrule[1.2pt]
        \end{tabular}
    }
    \vspace{-1em}
\end{table}

Multimodal large language models (MLLMs) extend LLMs by integrating visual inputs to support multimodal reasoning~\cite{chen2024mllm, reasoning, jain2024vcoder}. Typically, visual features are extracted by a vision encoder, aligned through a connector, and processed by the LLM~\cite{llava_origin, parekh2024concept}. Despite strong performance, MLLMs often produce factually incorrect or visually ungrounded outputs, undermining reliability~\cite{bai2024hallucination, chen2025perturbollava}. This issue is more severe than in unimodal LLMs~\cite{chen2024multi, jiang2024hallucination}, mainly due to modality imbalance~\cite{ma2024visual, he2024ma} and ineffective fusion~\cite{bellagente2023multifusion, ji2023map}. Recent studies attribute these failures to persistent misalignment between multimodal representations and human expectations, rather than model capacity~\cite{chen2024unified, liu2024aligning, hao2025hccm}.

A key bottleneck in addressing MLLM hallucinations lies in aligning model outputs with human preferences for factual consistency. Unlike knowledge-intensive pretraining~\cite{chang2024large, mckinzie2024mm1} and instruction tuning~\cite{chen2024coin, llava_origin}, recent methods typically leverage small-scale human preference data refined via reinforcement learning~\cite{yu2024rlhf, casperopen, feng2025rewardmap}. Direct preference optimization (DPO)~\cite{rafailov2023direct, DPO_MLLMs} has become a leading approach due to its simplicity and effectiveness, demonstrated in CHiP-DPO~\cite{CHIP} and OPA-DPO~\cite{OPA_DPO}. However, DPO's reliance on limited data can cause overfitting to superficial linguistic cues~\cite{setlur2024rl, CHIP}, leading to distributional rigidity and reduced adaptability to modality-specific semantics~\cite{ouali2024clip, song2024importance}. These limitations call for more adaptive alignment strategies that capture token-level variability and cross-modal dependencies for stable multimodal reasoning. Adversarial training has proven effective for improving robustness in vision-language pretraining~\cite{gan2020large, zhang2025towards}, yet its integration into preference optimization for hallucination mitigation remains largely unexplored.

To address these challenges, we propose a token-adaptive min-max alignment strategy with spectral regularization that enhances preference learning without relying on high-resource expert feedback (\eg, GPT-4V~\cite{achiam2023gpt}). Our spectral alignment loss operates in the frequency domain via FFT, encouraging global semantic consistency while tolerating the local distributional shifts induced by token perturbation. Using only a small public preference dataset, our method effectively mitigates hallucinations and consistently outperforms RL-based baselines across benchmarks. \cref{tb:comparison_PO} compares preference optimization methods in terms of data scale, supervision, and alignment.

\section{Conclusion}
\label{sec:conclusion}
This work reveals a fundamental limitation of static preference optimization: its tendency to latch onto superficial linguistic correlations rather than causally grounded visual evidence. We address this not by scaling data, but by reshaping the optimization landscape itself. TARS recasts preference learning as a min-max game in which the inner adversarial maximization over visual-agnostic tokens forces the model to disentangle causal visual signals from spurious textual cues, while a spectral alignment loss in the frequency domain provides a theoretically principled regularizer that preserves global semantic invariance under the resulting distributional shifts.
Empirically, using only 4.8k preference samples without expert feedback, TARS halves the hallucination rate (26.4\%\,$\to$\,13.2\%), reduces cognitive inconsistency by an order of magnitude (2.5\,$\to$\,0.4), and surpasses $5\times$ LLM-based augmentation on 28.8k samples while narrowing the gap with GPT-4o at a fraction of the model scale. These results establish that principled adversarial token perturbation can rival brute-force data scaling for robust multimodal preference alignment.

\setcounter{section}{6}

\section*{Appendix Overview}
This appendix provides additional details to support the main paper. It is organized as follows:

\begin{itemize}
    \item \textbf{Section~\ref{sec:implementation}} details model configurations, training settings for DPO and TARS, and token perturbation procedures.
    \item \textbf{Section~\ref{sec:algorithm}} presents the min-max optimization algorithm of TARS in pseudocode form.
    \item \textbf{Section~\ref{sec:extend_abalation}} includes extended ablation studies, covering perturbation magnitude, spectral regularization strength, a formal theoretical analysis of spectral alignment (energy dispersion and low-frequency masking bounds), spectral regularization alternatives, adaptive vs.\ random token selection, and min--max formulation analysis.
    \item \textbf{Section~\ref{sec:add_experiment}} reports additional benchmark results, including fine-grained AMBER hallucination metrics, extended MS-COCO CHAIR and multimodal understanding evaluations, comparison with data augmentation baselines, and comparison with retrieval-augmented DPO.
    \item \textbf{Section~\ref{sec:muffin_results}} presents results on Muffin-13B, demonstrating the generality of TARS across different MLLM architectures.
    \item \textbf{Section~\ref{sec:discussion}} discusses model behavior, including sensitivity and design insights.
    \item \textbf{Section~\ref{sec:examples}} showcases qualitative comparisons on representative examples.
\end{itemize}

\section{Implementation Details}
\label{sec:implementation}
\subsection{Base Model Setups}
We evaluate our method on LLaVA-v1.5~\cite{llava_origin} models with 7B and 13B parameters. LLaVA-v1.5 adopts Vicuna-7B/13B~\cite{chiang2023vicuna} as the language backbone and CLIP-ViT-L/14~\cite{radford2021learning} as the vision encoder. The vision encoder also serves as the similarity function $\mathcal{G}(\cdot)$ used in \cref{eq:perturbation} to compute alignment between visual inputs and text tokens. All experiments are conducted using greedy decoding with a temperature of 0 to ensure deterministic outputs and reproducibility.

\subsection{DPO Training Setups}
For fair comparison, DPO~\cite{wang2024mdpo}, CHiP~\cite{CHIP}, and TARS follow the same training protocol as described in CHiP~\cite{CHIP}. Specifically, we set the number of epochs to 3, learning rate to 5e-7, warmup ratio to 0.03, maximum sequence length to 2048, and gradient clipping threshold to 20.0. Notably, TARS requires no task-specific hyperparameter tuning and demonstrates generalization across different base models and datasets. All experiments are conducted on 8$\times$A100 GPUs (80GB). Each training run takes approximately 3.0 hours on LLaVA-v1.5-7B and 3.4 hours on LLaVA-v1.5-13B.

To generate perturbed inputs, we apply two token-level adversarial strategies: \textit{replace} and \textit{mask}. Both are guided by token similarity scores that estimate the alignment between each text token and the visual context. The similarity matrix is normalized into perturbation scores, such that tokens with lower alignment are more likely to be modified. In \texttt{replace} mode, these tokens are substituted with random vocabulary tokens. In \texttt{mask} mode, they are replaced with a special token such as \texttt{[MASK]}, \texttt{[UNK]}, or \texttt{[PAD]}, depending on tokenizer availability. Special tokens (\eg, \texttt{[BOS]}, \texttt{[EOS]}, \texttt{[PAD]}) are explicitly excluded from perturbation.

\subsection{Evaluation Benchmark Setups}
We evaluate TARS on four established hallucination benchmarks:

\noindent\textbf{1) AMBER}~\cite{wang2023amber} (Generative): A fine-grained benchmark for hallucination evaluation. Following prior works~\cite{OPA_DPO, CHIP}, we evaluate only the generative subset using the official codebase. Metrics include CHAIR~\cite{rohrbach2018object} for object hallucination detection, object coverage (Cover) for completeness, response-level hallucination rate (Hal-Rate), and alignment with human cognition (Cog).

\noindent\textbf{2) MMHal}~\cite{sun2023aligning} (Generative): A VQA benchmark with real-world scenarios, evaluated using GPT-4V feedback to measure overall scores and hallucination rates (Hal-Rate).

\noindent\textbf{3) OBJHal}~\cite{yu2024rlhf} (Generative): A benchmark evaluating hallucinations in image captioning. We report hallucination rates at the response level (CR$_s$) and object mention level (CR$_i$).

\noindent\textbf{4) POPE}~\cite{li2023evaluating} (Discriminative): A binary VQA benchmark designed to assess object hallucination through yes/no questions, using \texttt{popular}, \texttt{random}, and \texttt{adversarial} sampling strategies.

We follow the original evaluation settings and benchmark splits for AMBER, MMHal, and OBJHal as specified in their respective papers. For POPE, we construct a benchmark of 9,000 VQA pairs by sampling using the three strategies above.

For evaluation metrics, we adopt four response-level hallucination measures across different benchmarks: CHAIR~\cite{rohrbach2018object} for object hallucination detection, object coverage (Cover) for completeness measurement, response-level hallucination rate (Hal-Rate) for overall hallucination assessment, sentence-level hallucination rate (CR$_s$) for holistic response evaluation, and object mention-level hallucination rate (CR$_i$) for fine-grained object-level analysis.

For evaluation feedback collection, we employ the \texttt{en-core-web-lg} English NLP pipeline for AMBER to extract structured semantic cues as lightweight and reproducible evaluators. For MMHal and OBJHal, we utilize the expert GPT-4V model~\cite{hurst2024gpt} (\texttt{gpt-4-1106-vision-preview}) for feedback evaluation, following the established protocols.

\section{Algorithm Flowchart}
\label{sec:algorithm}
We present the full training procedure of TARS in Algorithm~\ref{alg:TARS}, which explicitly decomposes the learning process into two stages: a maximization phase that generates token-level perturbations based on visual relevance (\textit{\textcolor{red}{Max Part}}), and a minimization phase that optimizes the model with preference supervision (\textit{\textcolor{blue}{Min Part}}).
This min-max formulation allows TARS to effectively regularize overconfident preference patterns by injecting controlled distributional shifts during training.
The maximization step identifies visually agnostic tokens and perturbs them via masking or replacement, while the minimization step jointly optimizes a DPO loss and a frequency-domain alignment objective.
Overall, TARS effectively suppresses spurious token-visual correlations and significantly reduces hallucinations in multimodal preference optimization.

\begin{algorithm}[tb]
    \caption{\textit{TARS Training Procedure}}
    \label{alg:TARS}
    \begin{algorithmic}[1]
        \newcommand{\DPODATA}[1]{\item[]\hspace*{-1.55em}\textbf{Inputs:} #1}
        \newcommand{\ENCODERS}[1]{\item[]\hspace*{-1.55em}\textbf{Encoders:} #1}
        \newcommand{\HYPERPARAMS}[1]{\item[]\hspace*{-1.55em}\textbf{Hyperparameters:} #1}
        \newcommand{\RETURNPOLICY}[1]{\item[]\hspace*{-1.55em}\textbf{Learned Policy:} #1}
        \DPODATA{Trainable policy $\pi_{\theta}$, reference policy $\pi_{\text{ref}}$, and preference dataset $\mathcal{D} = \{x, q, y_w, y_r\}^N$.}
        \ENCODERS{Visual encoder $\mathcal{G}_v$; text encoder $\mathcal{G}_t$.}
        \HYPERPARAMS{DPO scaling $\alpha$, perturbation ratio $\omega$, frequency scaling $\beta$, loss weight $\lambda$.}

        \FOR{each epoch}
            \STATE Sample preference tuple $(x, q, y_w, y_r) \sim \mathcal{D}$.
            \STATE \textit{\textcolor{red}{Max Part:}}
            \STATE \hspace{0.7em} Compute token-level visual relevance:
            $P_i = \mathcal{G}_v(x) \cdot \mathcal{G}_t(q_i)^T$.
            \STATE \hspace{0.7em} Estimate model confidence margin:
            $\Delta P = \max_j P_j - \max_{k \neq j} P_k$.
            \STATE \hspace{0.7em} Determine adaptive perturbation budget:
            $N_t = \left\lfloor \omega \cdot \Delta P^{-1} \right\rfloor + 1$.
            \STATE \hspace{0.7em} Select visually agnostic tokens:
            $\mathcal{A} = \text{Top}_{N_t}(-P)$.
            \STATE \hspace{0.7em} Apply controlled perturbation to obtain $\varphi(q)$.
            \STATE \textit{\textcolor{blue}{Min Part:}}
            \STATE \hspace{0.7em} Compute the preference alignment loss $\mathcal{L}_{\text{DPO}}$ via DPO.
            \STATE \hspace{0.7em} Apply frequency-domain regularization $\mathcal{L}_{\text{freq}}$.
            \STATE \hspace{0.7em} Compute final objective:
            $\mathcal{L}_{\text{TARS}} = \mathcal{L}_{\text{DPO}} + \lambda \cdot \mathcal{L}_{\text{freq}}$.
            \STATE \hspace{0.7em} Update $\pi_\theta$ via gradient descent.
        \ENDFOR
        \RETURNPOLICY{Optimized policy $\pi_\theta^*$.}
    \end{algorithmic}
\end{algorithm}

\section{Extended Ablation Studies}
\label{sec:extend_abalation}

\subsection{Impact of Token-Level Perturbation Magnitude}
\label{sec:perturb-magnitude}
We vary the token-level perturbation ratio $\omega$ and report results in \cref{tb:magnitude} to investigate how perturbation strength affects model performance. In our method, \cref{eq:perturbation} governs the selection of tokens for perturbation based on their visual irrelevance. Specifically, we compute the similarity between visual features $\mathcal{G}_v(x)$ and text token embeddings $\mathcal{G}_t(q_i)$ to estimate token-level visual alignment. Tokens with the lowest scores are considered visual-agnostic and thus are most eligible for perturbation.

As shown in \cref{tb:magnitude}, moderate values of $\omega$ lead to optimal hallucination suppression across both AMBER and OBJHal. Excessively low or high perturbation strengths either under-regularize or destabilize training.
When $\omega$ is too small (\eg, $1\mathrm{e}{-4}$), the induced distributional shift is limited, resulting in marginal improvement over the baseline.
Conversely, overly large values (\eg, $5\mathrm{e}{-3}$ or $1\mathrm{e}{-2}$) introduce excessive perturbation into visual regions, disrupting the semantic coherence of inputs.
The best results are obtained at $\omega=1\mathrm{e}{-3}$, which achieves a balance between perturbation diversity and input integrity.

\begin{table}[tb]
\centering
    \caption{
    Ablation study on the effect of token-level perturbation magnitude $\omega$ in TARS. All experiments use LLaVA-v1.5-13B as the base model. Bold results indicate the best-performing configuration.}
    \label{tb:magnitude}
    \vspace{-1em}
    \resizebox{0.9\linewidth}{!}
    {
        \begin{tabular}{ccccccccccc}
        \toprule[1.2pt]
        \multirow{2}{*}{\shortstack{\textbf{Perturbation} \\ \textbf{Magnitude}}} &
        \multicolumn{4}{c}{\textbf{AMBER}} &
        \multicolumn{2}{c}{\textbf{MMHal}} &
        \multicolumn{2}{c}{\textbf{POPE}} &
        \multicolumn{2}{c}{\textbf{OBJHal}} \\
        \cmidrule[0.5pt](lr){2-5} \cmidrule[0.5pt](lr){6-7} \cmidrule[0.5pt](lr){8-9} \cmidrule[0.5pt](lr){10-11}
        & \textbf{CHAIR$\downarrow$} & \textbf{Cover$\uparrow$} & \textbf{Hal$\downarrow$} & \textbf{Cog$\downarrow$} & \textbf{Scr$\uparrow$} & \textbf{Hal$\downarrow$} & \textbf{Acc$\uparrow$} & \textbf{Pre$\uparrow$} & \textbf{CR$_\text{s}$$\downarrow$} & \textbf{CR$_\text{i}$$\downarrow$} \\
        \midrule
        \rowcolor{gray!10}
        \multicolumn{11}{l}{\textbf{Referenced Results}} \\
        LLaVA-7B&7.6&51.7&35.4&4.2&2.02&0.61&80.0&61.8&54.0&15.8 \\
        \midrule
        $\omega=1\mathrm{e}{-4}$ & 3.3 & 58.0 & 15.4 & 1.3 & 2.35 & 0.50 & 86.0 & 95.0 & 16.2 & 4.6 \\
        $\omega=3\mathrm{e}{-4}$ & 3.0 & 58.3 & 15.1 & 1.1 & 2.38 & 0.50 & 86.9 & 95.4 & 15.4 & 4.3 \\
        $\omega=5\mathrm{e}{-4}$ & 2.9 & 58.7 & 14.5 & 0.8 & 2.41 & 0.47 & 87.7 & 96.1 & 13.6 & 3.8 \\
        \rowcolor{orange!15}  $\omega=1\mathrm{e}{-3}$&\textbf{2.4}&\textbf{59.6}&\textbf{13.2}&\textbf{0.4}&\textbf{2.48}&\textbf{0.45}&\textbf{88.7}&\textbf{97.5}&\textbf{12.0}&\textbf{3.2} \\
        $\omega=5\mathrm{e}{-3}$ & 3.3 & 57.8 & 15.9 & 1.4 & 2.29 & 0.48 & 86.9 & 91.2 & 16.1 & 3.9 \\
        $\omega=1\mathrm{e}{-2}$ & 4.0 & 56.9 & 20.2 & 1.9 & 2.23 & 0.51 & 84.0 & 85.7 & 21.9 & 5.6 \\
        \bottomrule[1.2pt]
        \end{tabular}
    }
\end{table}

\subsection{Impact of Frequency Regularization}
To assess the contribution of spectral regularization, we conduct an ablation study by varying the frequency loss weight $\lambda$ in the TARS objective (\cref{eq:total_loss}). We evaluate $\lambda \in \{0.01, 0.02, 0.05, 0.10, 0.20, 0.50, 1.00\}$ and present results in \cref{tb:lambda_ablation}.

Performance improves steadily as $\lambda$ increases from 0.01 to 0.20, with hallucination rates decreasing and coverage improving. The best trade-off is achieved at $\lambda = 0.20$. Beyond this point, performance begins to degrade, particularly on MMHal and OBJHal. This trend suggests that overly aggressive regularization may constrain the model's ability to accommodate subtle semantic variations introduced by token-level perturbations.

\begin{table}[tb]
    \centering
    \caption{Ablation study on the effect of spectral alignment weight $\lambda$. All experiments are conducted using LLaVA-v1.5-13B. Bold numbers indicate the best across each metric.}
    \vspace{-1em}
    \label{tb:lambda_ablation}
    \resizebox{0.9\linewidth}{!}
    {
        \begin{tabular}{ccccccccccc}
        \toprule[1.2pt]
        \multirow{2}{*}{\shortstack{\textbf{Spectral} \\ \textbf{Coeff.}}} &
        \multicolumn{4}{c}{\textbf{AMBER}} &
        \multicolumn{2}{c}{\textbf{MMHal}} &
        \multicolumn{2}{c}{\textbf{POPE}} &
        \multicolumn{2}{c}{\textbf{OBJHal}} \\
        \cmidrule[0.5pt](lr){2-5} \cmidrule[0.5pt](lr){6-7} \cmidrule[0.5pt](lr){8-9} \cmidrule[0.5pt](lr){10-11}
        & \textbf{CHAIR$\downarrow$} & \textbf{Cover$\uparrow$} & \textbf{Hal$\downarrow$} & \textbf{Cog$\downarrow$} & \textbf{Scr$\uparrow$} & \textbf{Hal$\downarrow$} & \textbf{Acc$\uparrow$} & \textbf{Pre$\uparrow$} & \textbf{CR$_\text{s}$$\downarrow$} & \textbf{CR$_\text{i}$$\downarrow$} \\
        \midrule
        \rowcolor{gray!10}
        \multicolumn{11}{l}{\textbf{Referenced Results}} \\
        LLaVA-13B&6.7&52.1&32.5&3.5&2.39&0.53&74.6&55.2&50.0&14.5 \\
        \midrule
        $\lambda=0.01$ & 2.9 & 58.7 & 14.8 & 1.0 & 2.80 & 0.48 & 87.2 & 92.5 & 15.4 & 3.7 \\
        $\lambda=0.02$ & 2.7 & 59.0 & 14.1 & 0.8 & 2.83 & 0.46 & 87.5 & 93.2 & 14.8 & 3.4 \\
        $\lambda=0.05$ & 2.6 & 59.3 & 13.5 & 0.6 & 2.85 & 0.46 & 87.6 & 93.5 & 14.7 & 3.1 \\
        $\lambda=0.10$ & 2.4 & 59.6 & 13.2 & \textbf{0.4} & 2.88 & \textbf{0.45} & 88.2 & 94.3 & 13.2 & \textbf{2.9} \\
        \rowcolor{orange!15} $\lambda=0.20$ & \textbf{2.1} & \textbf{59.8} & \textbf{12.5} & 0.6 & \textbf{2.89} & \textbf{0.45} & \textbf{88.5} & \textbf{95.0} & \textbf{12.8} & 2.8 \\
        $\lambda=0.50$ & 2.6 & 59.0 & 13.9 & 0.9 & 2.86 & 0.46 & 87.8 & 92.4 & 14.4 & 3.5 \\
        $\lambda=1.00$ & 3.0 & 58.2 & 15.1 & 1.3 & 2.81 & 0.47 & 86.7 & 91.0 & 15.6 & 4.1 \\
        \bottomrule[1.2pt]
        \end{tabular}
    }
\end{table}

\subsection{Theoretical Analysis of Spectral Alignment}
\label{sec:spectral_theory}
We provide a formal justification for adopting frequency-domain alignment instead of spatial losses under token-level adversarial perturbations in TARS.

\noindent\textbf{Setup.}
Let a hidden-state sequence be $\mathbf{Z} = (\mathbf{z}_0, \dots, \mathbf{z}_{L-1}) \in \mathbb{R}^{L \times d}$. A single-token adversarial perturbation at position $m$ induces a corrupted sequence $\tilde{\mathbf{Z}}$:
\begin{equation}
\tilde{\mathbf{z}}_i = \mathbf{z}_i + \boldsymbol{\delta} \cdot \mathbb{I}(i = m),
\end{equation}
where $\boldsymbol{\delta} \in \mathbb{R}^d$ is bounded ($\|\boldsymbol{\delta}\| \le \epsilon$) and $\mathbb{I}(\cdot)$ is the indicator function.
The Discrete Fourier Transform (DFT) along the token axis maps $\mathbf{Z}$ into the frequency domain:
\begin{equation}
\hat{\mathbf{z}}_k = \sum_{i=0}^{L-1} \mathbf{z}_i \, e^{-j \frac{2\pi}{L} k i}, \quad k = 0, \dots, L{-}1,
\end{equation}
where $j$ denotes the imaginary unit.

\noindent\textbf{Proposition 1} (\textit{Spectral Energy Dispersion of Local Perturbation})\textbf{.}
\textit{Under a single-token perturbation at position $m$, the frequency-domain deviation is:}
\begin{equation}
\tilde{\hat{\mathbf{z}}}_k = \hat{\mathbf{z}}_k + \boldsymbol{\delta} \, e^{-j \frac{2\pi}{L} k m}.
\end{equation}
\textit{Taking the $\ell_2$ norm yields:}
\begin{equation}
\|\tilde{\hat{\mathbf{z}}}_k - \hat{\mathbf{z}}_k\|_2 = \|\boldsymbol{\delta}\|_2 \quad \forall\, k \in \{0, \dots, L{-}1\}.
\end{equation}
\textit{Proof.}
By linearity of the DFT, $\tilde{\hat{\mathbf{z}}}_k - \hat{\mathbf{z}}_k = \boldsymbol{\delta} \, e^{-j \frac{2\pi}{L} k m}$.
Since the complex exponential has unit modulus ($|e^{-j\theta}| = 1$), the $\ell_2$ norm of a real vector scaled by it is unchanged. \hfill$\blacksquare$

\noindent\textbf{Remark.}
Proposition~1 reveals that a concentrated spatial spike (single-token perturbation) acts as a Dirac-like impulse whose energy is \emph{uniformly dispersed} across all frequency bins $k$.

\noindent\textbf{Theorem 1} (\textit{Bounded Semantic Shift via Low-Frequency Masking})\textbf{.}
\textit{Let the spatial alignment loss be $\mathcal{L}_{\mathrm{spatial}} = \|\tilde{\mathbf{Z}} - \mathbf{Z}^{\mathrm{ref}}\|_F^2$. By Parseval's theorem,}
\begin{equation}
\mathcal{L}_{\mathrm{spatial}} = \frac{1}{L} \sum_{k=0}^{L-1} \|\tilde{\hat{\mathbf{z}}}_k - \hat{\mathbf{z}}^{\mathrm{ref}}_k\|_2^2.
\end{equation}
\textit{To decouple global semantics from local token noise, TARS introduces a low-frequency alignment loss with a binary mask $\mathcal{M}$ ($\mathcal{M}_k = 1$ for $k < K_{\mathrm{cut}}$, and $0$ otherwise):}
\begin{equation}
\mathcal{L}_{\mathrm{freq}} = \frac{1}{L} \sum_{k=0}^{L-1} \mathcal{M}_k \,\|\tilde{\hat{\mathbf{z}}}_k - \hat{\mathbf{z}}^{\mathrm{ref}}_k\|_2^2.
\end{equation}
\textit{Under the perturbation $\boldsymbol{\delta}$, the frequency loss is upper-bounded by:}
\begin{equation}
\mathcal{L}_{\mathrm{freq}} \le \frac{K_{\mathrm{cut}}}{L}\,\|\boldsymbol{\delta}\|^2 < \mathcal{L}_{\mathrm{spatial}}.
\end{equation}
\textit{Proof.}
Applying Proposition~1, each masked bin contributes at most $\|\boldsymbol{\delta}\|^2/L$ to $\mathcal{L}_{\mathrm{freq}}$. Summing over $K_{\mathrm{cut}} < L$ bins gives the bound. \hfill$\blacksquare$

\noindent\textbf{Interpretation for Min-Max Optimization.}
In the inner maximization of TARS, the model generates local adversarial shifts $\boldsymbol{\delta}$ to explore hallucination boundaries.
\begin{itemize}
    \item \textbf{Limitation of spatial loss.} Enforcing spatial alignment penalizes $\|\boldsymbol{\delta}\|^2$ entirely at the perturbed position $m$. This rigid positional constraint restricts the adversarial exploration space that the inner maximization is designed to exploit.
    \item \textbf{Advantage of spectral loss.} Because the dominant semantic structure of MLLM hidden states concentrates in the low-frequency modes (\ie, $\|\hat{\mathbf{z}}_k\|$ is substantially larger for small $k$), the relative distortion $\|\boldsymbol{\delta}\| / \|\hat{\mathbf{z}}_k\|$ in the masked low-frequency regime approaches zero. Thus, $\mathcal{L}_{\mathrm{freq}}$ acts as a robust invariant: it forces the global semantic envelope (low frequencies) to align with the non-hallucinated reference, while safely ignoring the high-frequency positional noise induced by the min-max perturbation.
\end{itemize}

\subsection{Analysis of Spectral Regularization Alternatives}
\label{sec:fft_ablation_study}
FFT-based regularization decomposes hidden representations into low- and high-frequency components along the token dimension.
\textbf{(1) Low-frequency alignment} preserves global semantic structure (\eg, topic, intent, and visual grounding) under token-level perturbations, since the dominant low-frequency modes capture the overall distributional statistics of the hidden-state sequence.
\textbf{(2) High-frequency constraint} mitigates overfitting to local lexical patterns and positional artifacts, improving robustness and generalization.
\textbf{(3) Why FFT over spatial metrics?}
Given hidden states $H=[h_1,\dots,h_T] \in \mathbb{R}^{T \times D}$, an $\ell_2$ loss $\|H - H'\|_2$ treats every positional deviation equally, amplifying position-dependent noise from our token perturbations. In contrast, the FFT representation $\mathcal{F}(H)$ aggregates information across all positions into frequency bins; a local perturbation at position $l$ contributes a bounded additive term $\propto e^{-2\pi i k l / L}$ to each frequency bin $k$, resulting in smooth spectral variation. As shown in \cref{tb:fft_ablation}, FFT-based regularization consistently outperforms $\ell_2$ and cosine baselines across all metrics, validating this design choice.
Additionally, \cref{tb:alignment_ablation} compares spectrum-based alignment (TARS) with token-level contrastive alignment under identical perturbation policies, showing that spectral alignment achieves substantially lower hallucination rates (13.2\% vs.\ 16.3\%) and higher coverage (59.6\% vs.\ 57.2\%).

\begin{table}[tb]
  \centering
  \caption{Ablation on spectral regularization alternatives. FFT-based regularization outperforms $\ell_2$ and cosine baselines.}
  \vspace{-1em}
  \label{tb:fft_ablation}
  \setlength{\tabcolsep}{4pt}
  \renewcommand{\arraystretch}{1.1}
  \resizebox{0.75\linewidth}{!}{
  \footnotesize{
    \begin{tabular}{lcccccc}
    \toprule
    Method & CHAIR$\downarrow$ & Cover$\uparrow$ & Hal-Rate$\downarrow$ & Cog$\downarrow$ & MMH$\uparrow$ & MMHal$\downarrow$ \\
    \midrule
    $\emptyset$ & 2.8 & 58.3 & 15.1 & 0.7 & 2.29 & 0.52 \\
    $\ell_2$  & 2.7 & 58.5 & 15.4 & 0.9 & 2.32 & 0.49 \\
    Cosine  & 2.5 & 58.8 & 14.5 & 0.7 & 2.34 & 0.49 \\
    \rowcolor{orange!20}
    FFT & \textbf{2.4} & \textbf{59.6} & \textbf{13.2} & \textbf{0.4} & \textbf{2.48} & \textbf{0.45} \\
    \bottomrule
    \end{tabular}
  }}
\end{table}

\subsection{Adaptive vs.\ Random Token Selection}
\label{sec:adaptive_selection}
A critical design choice in TARS is the \textit{adaptive} selection of perturbation targets based on cross-modal alignment scores, rather than random or uniform token selection. We validate this by comparing three strategies: (1) vision-only perturbation (perturbing tokens with \textit{high} visual relevance); (2) random perturbation (uniformly sampling tokens to perturb); and (3) our proposed visual-agnostic perturbation (selectively perturbing tokens with the \textit{lowest} cross-modal alignment).

As shown in \cref{tb:perturb_target}:
(1) Perturbing vision-related tokens degrades visual grounding, increasing CHAIR from 4.9 to 5.8 and Hal-Rate from 26.4\% to 31.6\%, confirming that visual tokens carry essential grounding information that must be preserved.
(2) Random perturbation provides modest improvements over DPO (Hal-Rate 28.7\% vs.\ 26.4\%) but falls far short of TARS (13.2\%), as it indiscriminately modifies both semantically critical and irrelevant tokens.
(3) Our adaptive visual-agnostic perturbation achieves the best results across all metrics, reducing Hal-Rate by 15.5 pp over random perturbation and providing a 2$\times$ improvement in Cog (0.4 vs.\ 3.1). These results \textbf{directly demonstrate that the advantage of TARS comes from the principled adaptive selection mechanism}, not merely from the act of perturbation itself.

\begin{table}[tb]
  \centering
  \caption{Comparison of perturbation target strategies. Visual-agnostic perturbation (TARS) consistently outperforms alternatives.}
  \vspace{-1em}
  \label{tb:perturb_target}
  \setlength{\tabcolsep}{4pt}
  \renewcommand{\arraystretch}{1.1}
  \resizebox{0.82\linewidth}{!}{
  \footnotesize{
    \begin{tabular}{lcccccc}
    \toprule
    Method & CHAIR$\downarrow$ & Cover$\uparrow$ & Hal-Rate$\downarrow$ & Cog$\downarrow$ & MMH$\uparrow$ & MMHal$\downarrow$ \\
    \midrule
    LLaVA-v1.5-7B & 7.6 & 51.7 & 35.4 & 4.2 & 2.02 & 0.61 \\
    + DPO & 4.9 & 56.6 & 26.4 & 2.5 & 2.19 & 0.61 \\
    + Vision-only Perturb. & 5.8 & 55.2 & 31.6 & 3.8 & 2.12 & 0.66 \\
    + Random Perturb. & 4.4 & 57.8 & 28.7 & 3.1 & 2.30 & 0.64 \\
    \rowcolor{orange!20}
    + TARS (ours) & \textbf{2.4} & \textbf{59.6} & \textbf{13.2} & \textbf{0.4} & \textbf{2.48} & \textbf{0.45} \\
    \bottomrule
    \end{tabular}
  }}
\end{table}

\subsection{Min--Max Formulation Analysis}
\label{sec:minmax_analysis}
A key design choice in TARS is the adversarial min--max formulation (\cref{eq:minmax_dpo}), which replaces the static preference optimization of standard DPO with a dynamic two-stage procedure. The full empirical analysis of training dynamics is presented in the main text (\cref{sec:training_dynamics}, \cref{fig:minmax_demo}). Here we provide additional protocol details.

We train all three variants (standard DPO, DPO with random perturbation (RandPert), and TARS (Min-Max)) on LLaVA-v1.5-7B with identical hyperparameters using 4.8k preference samples over 20 epochs. All reported loss values are normalized to $[0,1]$ to enable fair cross-method comparison. The test set is evaluated under two conditions: \textit{clean} (original held-out samples) and \textit{robust} (token-perturbed held-out samples), isolating generalization from robustness. The results confirm that: (1)~DPO achieves the lowest training loss ($\sim$0.2) but the worst robust test loss (0.65--0.75), exhibiting severe overfitting; (2)~RandPert yields intermediate but highly unstable convergence; and (3)~the Min-Max formulation achieves the best test performance on \emph{both} clean ($<$0.3) and robust ($<$0.5) splits, acting as an implicit regularizer that prevents premature convergence to spurious patterns.

\section{Additional Experimental Results}
\label{sec:add_experiment}

\subsection{AMBER Generative and Discriminative Metrics}
\label{sec:amber_extended}
We present extended results on the AMBER benchmark in \cref{tb:amber_generative_discriminative}, evaluating hallucination performance from both generative and discriminative perspectives.

The left portion of the table reports generative metrics, including CHAIR, Coverage, Hallucination Rate, and Cognitive Score. TARS achieves substantial improvements across all, reducing hallucination by over 13 points compared to DPO, and significantly improving image grounding as reflected by higher coverage and cognitive consistency.

Beyond generative evaluation, we further introduce fine-grained discriminative metrics that assess hallucination across four categories: Existence, Relation, Attribute, and Action. As shown in the right half of the table, TARS consistently outperforms both DPO and the LLaVA baseline in all dimensions.

\begin{table}[tb]
    \centering
    \caption{
    Comparison of generative and fine-grained discriminative hallucination metrics on the AMBER benchmark.}
    \vspace{-1em}
    \label{tb:amber_generative_discriminative}
    \resizebox{0.8\linewidth}{!}
    {
        \begin{tabular}{lcccccccc}
        \toprule
        \multirow{2}{*}{\textbf{Algorithm}} &
        \multicolumn{4}{c}{\textbf{Generative}} &
        \multicolumn{4}{c}{\textbf{Discriminative}}\\
        \cmidrule(lr){2-5} \cmidrule(lr){6-9}
        & \textbf{CHAIR$\downarrow$} & \textbf{Cover$\uparrow$} & \textbf{Hal$\downarrow$} & \textbf{Cog$\downarrow$} & \textbf{Exist.$\uparrow$} & \textbf{Rel.$\uparrow$} & \textbf{Attr.$\uparrow$} & \textbf{Act.$\uparrow$}\\
        \midrule
        \textbf{LLaVA-7B}~\cite{llava_origin}&7.9&54.7&37.1&3.2&82.9&58.6&65.6&70.1\\
        \hspace{3pt} + \textit{DPO}&4.9&56.6&26.4&2.5&87.1&59.7&74.6&79.4\\
        \rowcolor{orange!15}
        \hspace{3pt} + \textit{TARS}&\textbf{2.4}&\textbf{59.6}&\textbf{13.2}&\textbf{0.4}&\textbf{95.3}&\textbf{62.8}&\textbf{78.6}&\textbf{86.5}\\
        \midrule
        \textbf{LLaVA-13B}~\cite{llava_origin}&6.7&52.1&32.5&3.5&94.1&45.5&70.1&76.2\\
        \hspace{3pt} + \textit{DPO}&4.1&56.7&54.3&2.2&95.0&58.8&73.1&81.5\\
        \rowcolor{orange!15}
        \hspace{3pt} + \textit{TARS}&\textbf{2.1}&\textbf{59.8}&\textbf{12.5}&\textbf{0.6}&\textbf{98.9}&\textbf{67.0}&\textbf{82.0}&\textbf{86.6}\\
        \bottomrule
        \end{tabular}
    }
\end{table}

\subsection{Extended Hallucination Benchmarks}
\label{sec:extended_benchmarks}
We provide extended evaluation on two additional benchmark suites, MS-COCO~\cite{lin2014microsoft} CHAIR and multimodal understanding benchmarks, to demonstrate that TARS improves hallucination suppression without degrading broader multimodal reasoning.

\noindent\textbf{MS-COCO CHAIR (\cref{tb:mscoco_chair}).}
We evaluate on the standard MS-COCO~\cite{lin2014microsoft} CHAIR benchmark using three evaluation protocols:
CHAIR$^{-}$ (negative-sample filtering), CHAIR$^{1}$ (single-caption), and CHAIR$^{2}$ (multi-caption), each reported at both instance-level ($I$) and sentence-level ($S$).
TARS reduces CHAIR$_I^{-}$ from 10.5 (LLaVA) to 5.9 (--4.6~pp) and CHAIR$_S^{-}$ from 32.7 to 17.2 (--15.5~pp), outperforming DPO across all six metrics.
The improvements are most pronounced on the harder CHAIR$^2$ protocol (multi-caption), where TARS achieves 38.5\% instance-level vs.\ 49.8\% for DPO, a 22.7\% relative reduction.
These results confirm that TARS effectively mitigates object hallucination across diverse captioning settings.

\noindent\textbf{LLaVA-Bench and SeedBench (\cref{tb:multimodal_bench}).}
To verify that hallucination suppression does not impair general multimodal capabilities, we evaluate on LLaVA-Bench~\cite{llava_origin} (conversation, detail, and reasoning sub-tasks) and SeedBench~\cite{li2023seedbench} (multi-choice visual reasoning).
TARS improves over both the base LLaVA and DPO across all sub-tasks:
the overall LLaVA-Bench score increases from 62.9 (LLaVA) to 67.2 (+4.3), and SeedBench accuracy rises from 32.4\% to 38.7\% (+6.3~pp). Notably, the largest gain appears in visual reasoning (+3.9 over LLaVA, +2.3 over DPO), indicating that TARS strengthens the model's ability to perform multi-step inference grounded in visual evidence.

\begin{table}[tb]
  \centering
  \caption{Extended MS-COCO CHAIR evaluation across multiple settings.}
  \vspace{-1em}
  \label{tb:mscoco_chair}
  \setlength{\tabcolsep}{4pt}
  \renewcommand{\arraystretch}{1.1}
  \resizebox{0.85\linewidth}{!}{
  \footnotesize{
    \begin{tabular}{lcccccc}
        \toprule
        MS-COCO CHAIR
        & CHAIR$_I^{-}$ & CHAIR$_S^{-}$
        & CHAIR$^{1}_{I}$ & CHAIR$^{1}_{S}$
        & CHAIR$^{2}_{I}$ & CHAIR$^{2}_{S}$ \\
        \midrule
        LLaVA-v1.5-7B
        & 10.5 & 32.7
        & 18.8 & 62.7
        & 64.3 & 90.7 \\
        DPO
        & 7.2 & 21.4
        & 13.6 & 41.9
        & 49.8 & 72.3 \\ \rowcolor{orange!20}
        TARS (ours)
        & \textbf{5.9} & \textbf{17.2}
        & \textbf{10.8} & \textbf{33.6}
        & \textbf{38.5} & \textbf{57.9} \\
        \bottomrule
    \end{tabular}
  }}
\end{table}

\begin{table}[tb]
  \centering
  \caption{Evaluation on LLaVA-Bench and SeedBench, showing TARS improves multimodal understanding.}
  \vspace{-1em}
  \label{tb:multimodal_bench}
  \setlength{\tabcolsep}{4pt}
  \renewcommand{\arraystretch}{1.1}
  \resizebox{0.78\linewidth}{!}{
  \footnotesize{
    \begin{tabular}{lccccc}
    \toprule
    Method & Conv. & Detail & Reasoning & Overall & SeedBench Acc. \\
    \midrule
    LLaVA-v1.5-7B & 61.8 & 53.6 & 71.2 & 62.9 & 32.4 \\
    DPO           & 63.5 & 55.4 & 72.8 & 64.6 & 37.1 \\
    \rowcolor{orange!20}
    TARS (ours)   & \textbf{65.9} & \textbf{57.8} & \textbf{75.1} & \textbf{67.2} & \textbf{38.7} \\
    \bottomrule
    \end{tabular}
  }}
\end{table}

\subsection{Comparison with Data Augmentation Baselines}
\label{sec:augmentation_comparison}
A natural question is whether the gains of TARS can be replicated by simply augmenting the preference training data. We conduct a systematic comparison against two families of data augmentation strategies applied on top of standard DPO:

\noindent\textbf{(1) Paraphrasing.} We use GPT-3.5-Turbo to back-translate each preferred response through Chinese and French, then paraphrase back to English. This yields one additional variant per preference pair, preserving semantic content while diversifying surface forms. The augmented pairs are mixed with the original 4.8k samples during DPO training.

\noindent\textbf{(2) LLM-based augmentation.} We prompt an instruction-tuned LLM (Vicuna-13B~\cite{chiang2023vicuna}) to generate new preferred responses conditioned on the same image-question pairs, using the original preferred response as a reference. We experiment with two scales: $1\times$ (one new response per pair, yielding 9.6k total) and $5\times$ (five new responses per pair, yielding 28.8k total). Generated responses are filtered by CLIP score to remove low-quality samples before mixing with the original data.

\noindent\textbf{Protocol.} All models use LLaVA-v1.5-7B as the base and train with identical hyperparameters (learning rate, batch size, number of epochs) to isolate the effect of data composition. The only variable is the training data: original 4.8k samples vs.\ augmented sets.

\noindent\textbf{Analysis.}
As shown in \cref{tb:augmentation_appendix}, augmentation provides incremental improvements: paraphrasing reduces Hal-Rate from 26.4\% to 24.8\% (--1.6~pp), and $5\times$ LLM augmentation further lowers it to 16.0\% (--10.4~pp). However, even the most aggressive augmentation ($5\times$, using $6\times$ more data) still underperforms TARS on every metric (Hal-Rate 16.0\% vs.\ 13.2\%, Cog 1.2 vs.\ 0.4). This gap demonstrates that TARS provides a fundamentally different and stronger form of robustness that cannot be recovered through data scaling alone.

Crucially, TARS and augmentation are complementary: combining TARS with $5\times$ LLM augmentation achieves the best overall performance (CHAIR~2.1, Cover~60.8\%, Hal-Rate~12.3\%). This suggests that augmentation diversifies the preference distribution while TARS sharpens the model's sensitivity to token-level hallucination cues, and the two mechanisms address orthogonal failure modes.

\begin{table}[tb]
  \centering
  \caption{Comparison of TARS against data augmentation baselines across four hallucination benchmarks. \textit{Paraphrasing} denotes back-translation-based response diversification; \textit{LLM Aug.} denotes LLM-generated preferred responses, where $1\times$ and $5\times$ indicate the expansion factor relative to the original 4.8k preference set. All methods build on standard DPO. TARS is complementary to augmentation strategies.}
  \vspace{-1em}
  \label{tb:augmentation_appendix}
  \setlength{\tabcolsep}{3pt}
  \renewcommand{\arraystretch}{1.1}
  \resizebox{\linewidth}{!}{
  \footnotesize{
    \begin{tabular}{lcccccccccc}
        \toprule
        \multirow{2}{*}{\textbf{Method}} &
        \multicolumn{4}{c}{\textbf{AMBER}} &
        \multicolumn{2}{c}{\textbf{MMHal}} &
        \multicolumn{2}{c}{\textbf{POPE}} &
        \multicolumn{2}{c}{\textbf{OBJHal}} \\
        \cmidrule[0.5pt](lr){2-5} \cmidrule[0.5pt](lr){6-7} \cmidrule[0.5pt](lr){8-9} \cmidrule[0.5pt](lr){10-11}
        & CHAIR$\downarrow$ & Cover$\uparrow$ & Hal-Rate$\downarrow$ & Cog$\downarrow$
        & Score$\uparrow$ & Hal-Rate$\downarrow$
        & Acc$\uparrow$ & Pre$\uparrow$
        & CR$_\text{s}$$\downarrow$ & CR$_\text{i}$$\downarrow$ \\
        \midrule
        LLaVA-v1.5-7B-DPO
        & 4.9 & 56.6 & 26.4 & 2.5 & 2.19 & 0.61 & 87.8 & 82.0 & 14.0 & 5.0 \\
        + Paraphrasing
        & 4.5 & 57.9 & 24.8 & 2.3 & 2.24 & 0.58 & 87.9 & 83.5 & 13.6 & 4.8 \\
        + LLM Aug.\ (1$\times$)
        & 4.1 & 58.8 & 22.9 & 2.0 & 2.30 & 0.55 & 88.1 & 85.8 & 13.2 & 4.5 \\
        + LLM Aug.\ (5$\times$)
        & 3.1 & 59.3 & 16.0 & 1.2 & 2.38 & 0.50 & 88.3 & 90.4 & 12.6 & 4.0 \\
        \rowcolor{orange!20}
        + TARS
        & {2.4} & {59.6} & {13.2} & {0.4} & {2.48} & {0.45} & {88.7} & {97.5} & {12.0} & {3.2} \\
        \rowcolor{orange!20}
        + TARS \& Paraphrasing
        & 2.3 & 60.1 & 12.8 & 0.4 & 2.50 & 0.44 & 88.8 & 97.6 & 11.5 & 3.1 \\
        \rowcolor{orange!20}
        + TARS \& LLM Aug.\ (1$\times$)
        & 2.2 & 60.5 & 12.5 & 0.3 & 2.53 & 0.43 & 88.9 & 97.7 & 11.3 & 3.1 \\
        \rowcolor{orange!20}
        + TARS \& LLM Aug.\ (5$\times$)
        & \textbf{2.1} & \textbf{60.8} & \textbf{12.3} & \textbf{0.2} & \textbf{2.55} & \textbf{0.42} & \textbf{89.1} & \textbf{97.8} & \textbf{10.6} & \textbf{2.9} \\
        \bottomrule
    \end{tabular}
  }}
\end{table}

\subsection{Comparison with Retrieval-Augmented DPO}
\label{sec:realign_comparison}
RE-ALIGN~\cite{xing2025realign} augments standard DPO by injecting retrieved visual evidence, specifically relevant image patches and captions retrieved via CLIP, into the preference optimization pipeline. This approach enhances the factual grounding of preference pairs by providing the model with additional visual context during training. In contrast, TARS improves robustness through an entirely orthogonal mechanism: adversarial token-level perturbation within a min--max framework, without modifying the input data or retrieval pipeline.

\noindent\textbf{Experimental setup.} We compare four configurations on LLaVA-v1.5-7B using the same 4.8k preference samples: base LLaVA, standard DPO, DPO + RE-ALIGN, and TARS + RE-ALIGN. For RE-ALIGN, we follow the original implementation using CLIP-ViT-L/14 for retrieval and inject the top-3 retrieved evidence snippets.

\noindent\textbf{Analysis.}
As shown in \cref{tb:realign}, RE-ALIGN alone provides strong improvements over standard DPO, reducing Hal-Rate from 26.4\% to 16.8\% (--9.6~pp) and improving object coverage from 56.6\% to 58.9\% (+2.3~pp). However, combining TARS with RE-ALIGN yields substantial further gains across all six metrics: Hal-Rate drops to 12.5\% (--4.3~pp below RE-ALIGN alone), CHAIR decreases from 3.6 to 2.2, and MMH improves from 2.31 to 2.57.

The complementarity of the two methods can be understood through their distinct mechanisms: RE-ALIGN enriches the \emph{input space} by providing additional visual evidence, reducing hallucinations caused by insufficient grounding information. TARS, on the other hand, strengthens the \emph{optimization process} by exposing the model to adversarial token perturbations during training, reducing hallucinations caused by fragile token-level associations. Together, they address both data-side and algorithm-side sources of hallucination, achieving the best performance across all metrics.

\begin{table}[tb]
  \centering
  \caption{Complementarity of TARS and RE-ALIGN. Combining both methods achieves the best performance.}
  \vspace{-1em}
  \label{tb:realign}
  \setlength{\tabcolsep}{4pt}
  \renewcommand{\arraystretch}{1.1}
  \resizebox{0.82\linewidth}{!}{
  \footnotesize{
    \begin{tabular}{lcccccc}
    \toprule
    Method & CHAIR$\downarrow$ & Cover$\uparrow$ & Hal-Rate$\downarrow$ & Cog$\downarrow$ & MMH$\uparrow$ & MMHal$\downarrow$ \\
    \midrule
    LLaVA-v1.5-7B & 7.6 & 51.7 & 35.4 & 4.2 & 2.02 & 0.61 \\
    + DPO & 4.9 & 56.6 & 26.4 & 2.5 & 2.19 & 0.61 \\
    + RE-ALIGN & 3.6 & 58.9 & 16.8 & 1.3 & 2.31 & 0.52 \\
    \rowcolor{orange!20}
    + TARS \& RE-ALIGN & \textbf{2.2} & \textbf{60.4} & \textbf{12.5} & \textbf{0.4} & \textbf{2.57} & \textbf{0.43} \\
    \bottomrule
    \end{tabular}
  }}
\end{table}

\section{Additional Results on Muffin}
\label{sec:muffin_results}

We further validate TARS on the Muffin-13B architecture (\cref{tb:sota_muffin}). Consistent with our findings on LLaVA, both perturbation strategies yield substantial improvements over DPO and CHiP-DPO. TARS with token masking achieves the strongest overall performance, while synonym replacement remains competitive. These results confirm the versatility of our approach across different MLLM backbones.

\begin{table}[tb]
    \caption{Comparison of hallucination benchmarks on Muffin-13B. \textbf{Bold} denotes the best, and \underline{underlined} denotes the second-best.}
    \label{tb:sota_muffin}
    \vspace{-1em}
    \resizebox{0.9\linewidth}{!}
    {
        \begin{tabular}{lcccccccccc}
        \toprule
        \multirow{2}{*}{\textbf{Algorithm}} &
        \multicolumn{4}{c}{\textbf{AMBER}} &
        \multicolumn{2}{c}{\textbf{MMHal}} &
        \multicolumn{2}{c}{\textbf{POPE}} &
        \multicolumn{2}{c}{\textbf{OBJHal}} \\
        \cmidrule(lr){2-5} \cmidrule(lr){6-7} \cmidrule(lr){8-9} \cmidrule(lr){10-11}
        & \textbf{CHAIR$\downarrow$} & \textbf{Cover$\uparrow$} & \textbf{Hal$\downarrow$} & \textbf{Cog$\downarrow$} & \textbf{Scr$\uparrow$} & \textbf{Hal$\downarrow$} & \textbf{Acc$\uparrow$} & \textbf{Pre$\uparrow$} & \textbf{CR$_\text{s}$$\downarrow$} & \textbf{CR$_\text{i}$$\downarrow$} \\
        \midrule
        \textbf{Muffin-13B}~\cite{yu2023reformulating}
        &7.5&45.7&34.6&3.4&2.27&0.58&83.0&80.7&47.3&15.2 \\
        \hspace{3pt} + \textit{RLHF}~\cite{RLHF}
        &7.1&45.2&37.1&3.5&2.12&0.64&84.0&79.8&45.5&12.7 \\
        \hspace{3pt} + \textit{DPO}~\cite{li2024multi}
        &6.0&46.4&29.6&2.8&2.45&0.55&83.7&81.2&43.8&13.9 \\
        \hspace{3pt} + \textit{CHiP-DPO}~\cite{CHIP}
        &4.8&48.2&18.9&1.7&2.70&0.47&84.5&82.1&35.2&11.5 \\
        \rowcolor{orange!15} \hspace{3pt} + TARS~(Mask)
        &\textbf{3.6}&\textbf{49.5}&\textbf{16.2}&\textbf{1.4}&\underline{2.75}&\textbf{0.41}&\textbf{87.4}&\textbf{84.9}&\textbf{28.7}&\textbf{8.2} \\
        \rowcolor{orange!15} \hspace{3pt} + TARS~(Replace)
        &\underline{4.0}&\underline{48.9}&\underline{16.5}&\underline{1.5}&\textbf{2.76}&\underline{0.43}&\underline{86.5}&\underline{83.8}&\underline{29.3}&\underline{8.8} \\
        \bottomrule
        \end{tabular}
    }
\end{table}

\section{Discussions and Insights}
\label{sec:discussion}
\subsection{Why TARS Outperforms DPO: Beyond Numbers}
While TARS consistently outperforms standard DPO across hallucination benchmarks, its effectiveness stems not only from empirical gains, but from the design principles that enable better preference alignment under uncertainty.

\noindent\textbf{Token-level perturbation enhances alignment robustness.}
DPO relies on static textual inputs, making it susceptible to overfitting on superficial linguistic patterns. TARS addresses this issue by introducing controlled perturbations on visually agnostic tokens. These perturbations simulate semantically equivalent variations, exposing the model to distributional shifts during training. As a result, the learned policy becomes more robust to alignment uncertainty.

\noindent\textbf{Visual-agnostic targeting preserves grounding fidelity.}
Unlike random or uniform perturbation strategies, TARS selectively perturbs tokens with low cross-modal relevance. This design ensures that semantic shifts are injected without disrupting the causal connection between image and text, resulting in faithful responses that remain sensitive to visual semantics while being resilient to linguistic noise.

\noindent\textbf{Spectral alignment encourages semantic consistency.}
TARS introduces a spectral regularizer that aligns representations in the frequency domain. This global constraint allows for flexible modifications while maintaining semantic coherence at the sequence level, discouraging the model from latching onto spurious token-level correlations.

\begin{table}[tb]
    \centering
    \caption{Comparison of spectrum-based alignment (TARS) versus token-level contrastive alignment. Both models use identical perturbation policies.}
    \label{tb:alignment_ablation}
    \vspace{-1em}
    \resizebox{0.85\linewidth}{!}
    {
        \begin{tabular}{ccccccccccc}
        \toprule
        \multirow{2}{*}{\textbf{Strategy}} &
        \multicolumn{4}{c}{\textbf{AMBER}} &
        \multicolumn{2}{c}{\textbf{MMHal}} &
        \multicolumn{2}{c}{\textbf{POPE}} &
        \multicolumn{2}{c}{\textbf{OBJHal}} \\
        \cmidrule(lr){2-5} \cmidrule(lr){6-7} \cmidrule(lr){8-9} \cmidrule(lr){10-11}
        & \textbf{CH$\downarrow$} & \textbf{Cov$\uparrow$} & \textbf{Hal$\downarrow$} & \textbf{Cog$\downarrow$} & \textbf{Scr$\uparrow$} & \textbf{Hal$\downarrow$} & \textbf{Acc$\uparrow$} & \textbf{Pre$\uparrow$} & \textbf{CR$_\text{s}$$\downarrow$} & \textbf{CR$_\text{i}$$\downarrow$} \\
        \midrule
        Token-level & 3.4 & 57.2 & 16.3 & 1.4 & 2.36 & 0.49 & 87.1 & 93.3 & 15.8 & 4.9 \\
        \rowcolor{orange!15}
        Spectrum (TARS) & \textbf{2.4} & \textbf{59.6} & \textbf{13.2} & \textbf{0.4} & \textbf{2.48} & \textbf{0.45} & \textbf{88.7} & \textbf{97.5} & \textbf{12.0} & \textbf{3.2} \\
        \bottomrule
        \end{tabular}
    }
\end{table}

\subsection{Limitations}
In this work, we adopt two simple perturbation strategies: token masking and synonym replacement. These methods are chosen for their clarity, efficiency, and ease of interpretation. However, their simplicity may limit the generality and flexibility of the approach. Future work could explore adaptive or data-driven perturbation mechanisms~\cite{bai2026dice} that better balance semantic preservation with distributional shift. Additionally, the current token selection strategy, based on cross-modal similarity heuristics, could be enhanced by learning-based relevance estimation or causal attribution techniques.

\section{Qualitative Examples}
\label{sec:examples}
\begin{figure}[tb]
    \centering
    \begin{tabular}{c}
        \includegraphics[width=0.9\linewidth]{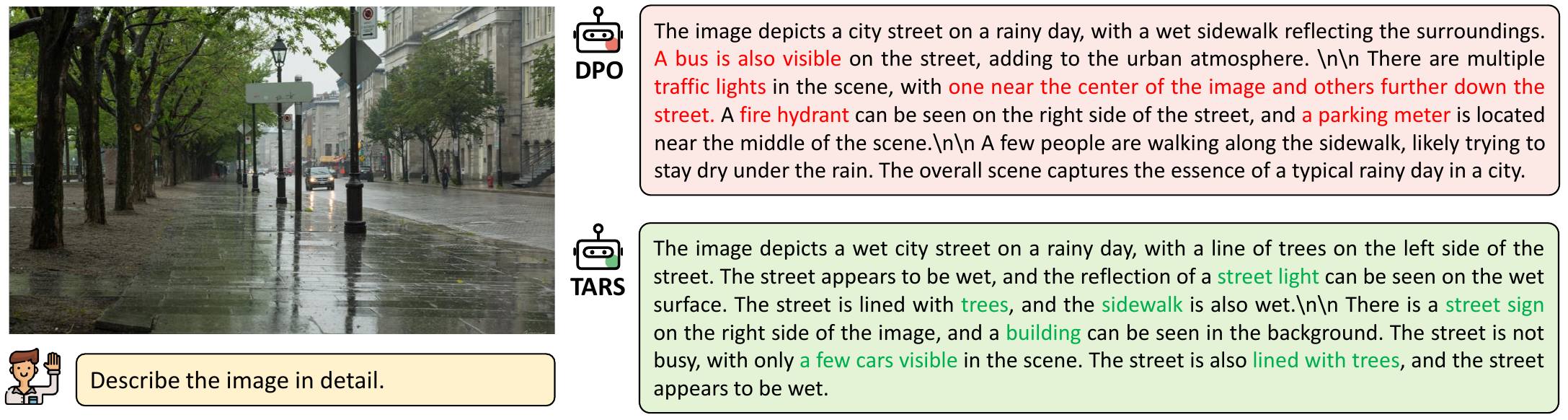}  \\
        (a)  \\
        \includegraphics[width=0.9\linewidth]{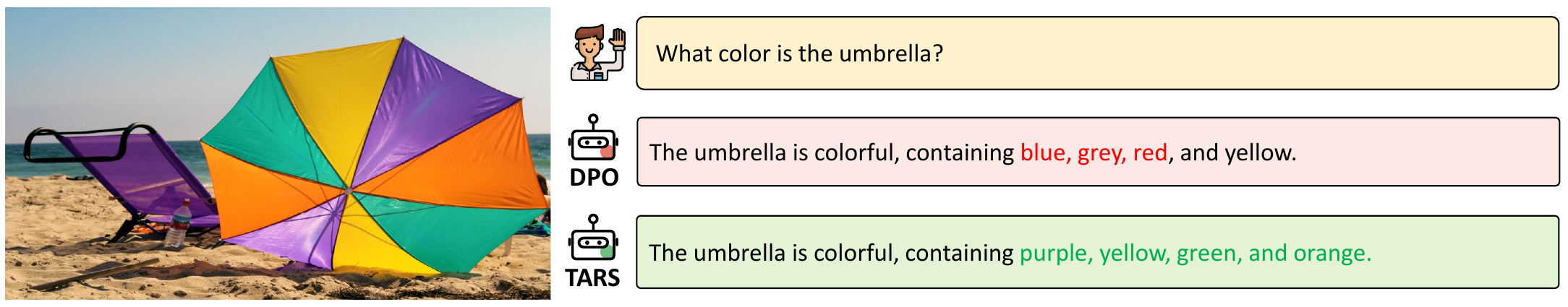} \\
        (b)  \\
        \includegraphics[width=0.9\linewidth]{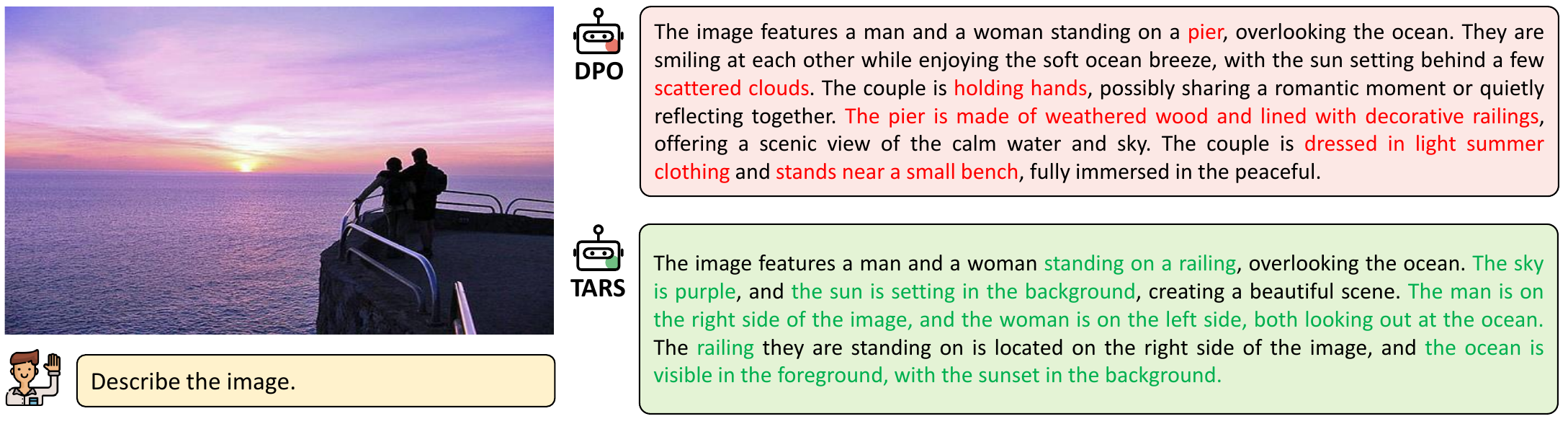}  \\
        (c)  \\
        \includegraphics[width=0.9\linewidth]{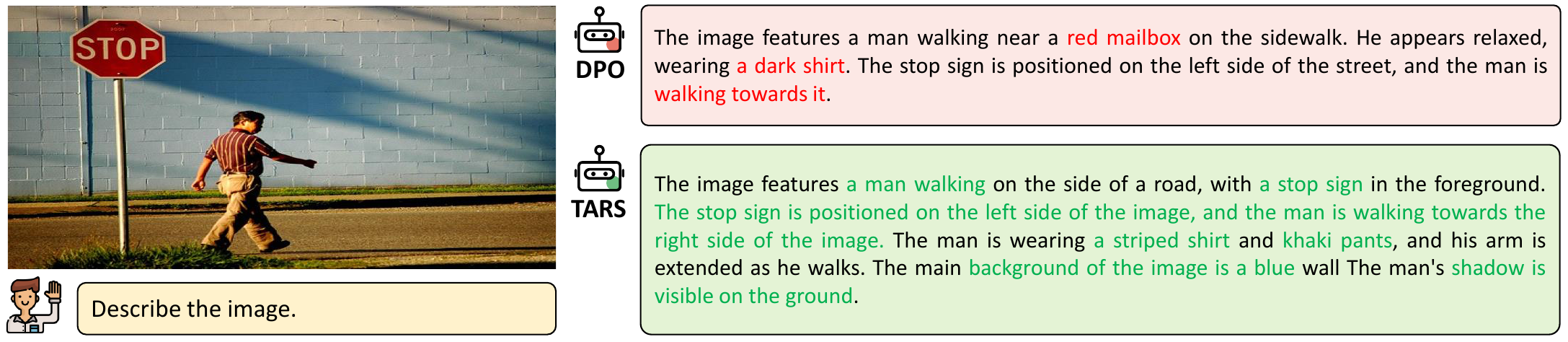}  \\
        (d)  \\
        \includegraphics[width=0.9\linewidth]{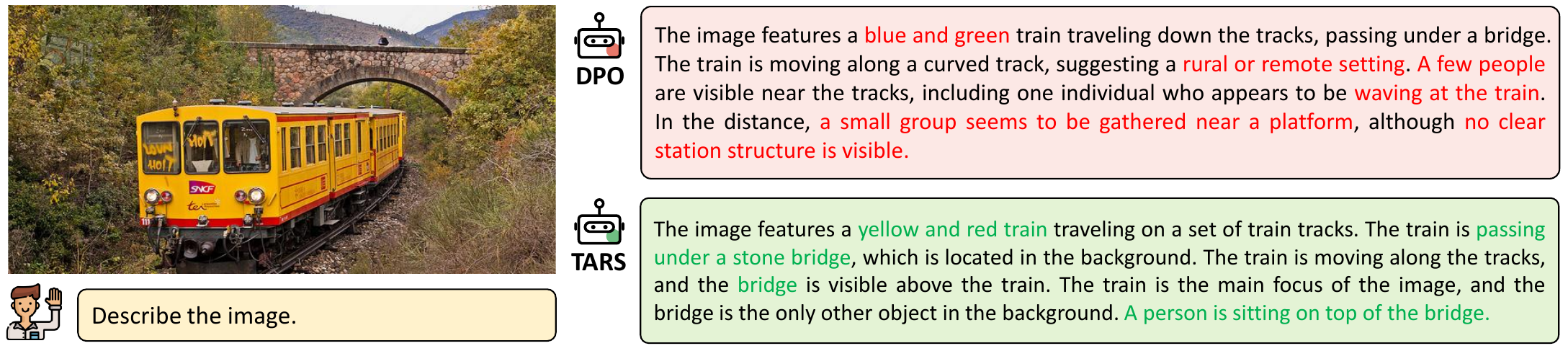}  \\
        (e)  \\
    \end{tabular}
    \caption{
    Qualitative comparisons between DPO and TARS across five diverse image-prompt pairs, denoted as (a)--(e).
    Hallucinated content is highlighted in \textcolor{red}{red}, while accurate visual grounding is marked in \textcolor{darkgreen}{green}.
    TARS consistently produces more faithful and informative responses.
    }
    \label{fig:demo_examples}
\end{figure}

We provide qualitative comparisons between standard DPO and our proposed TARS in \cref{fig:demo_examples}, across diverse image-question pairs.
TARS consistently demonstrates improved grounding and hallucination suppression, outperforming traditional DPO in several key aspects:

\noindent\textbf{Reduced hallucination via improved visual grounding.} Compared to DPO, TARS produces responses that more accurately reflect the image content. In all cases, DPO introduces visual details not present in the input, while TARS remains faithful to the scene.

\noindent\textbf{No degradation in response completeness.} TARS maintains response richness without sacrificing length or informativeness.

\noindent\textbf{Better fine-grained grounding.} TARS exhibits improved alignment at the fine-grained level, such as object color and count.

\noindent\textbf{Enhanced descriptive richness.} TARS responses incorporate more relevant visual details, indicating stronger cross-modal alignment.

\clearpage

\bibliographystyle{splncs04}
\bibliography{main}

\end{document}